\documentclass{IEEEtran}
\usepackage{fancyhdr}
\usepackage{times}
\usepackage{multicol}
\usepackage{graphics}
\usepackage{svg}
\usepackage{amsmath,amssymb,bm}
\usepackage{setspace}
\usepackage[Symbol]{upgreek}
\usepackage{graphicx}
\usepackage{floatrow} 
\usepackage{multirow}
\usepackage{array}
\usepackage{enumerate}
\usepackage{sidecap}
\usepackage[all]{xy}
\usepackage{authblk}
\usepackage{comment}
\usepackage{changepage}   
\usepackage{wrapfig}
\usepackage{subfig}
\usepackage{caption}
\usepackage{url}
\usepackage{tabularx,pbox,booktabs}
\usepackage{flushend}
\usepackage{anyfontsize}
\usepackage{float}
\usepackage{todonotes}
\usepackage[percent]{overpic}
\graphicspath{{./pics/}}
\usepackage{amssymb}
\usepackage{dblfloatfix}    
\usepackage{booktabs}
\usepackage{amsfonts}
\usepackage[T1]{fontenc}

\usepackage{mathabx}
\usepackage{upgreek}
\usepackage{algpseudocode}
\usepackage{algorithm}
\usepackage{amsthm}
\makeatletter
\let\NAT@parse\undefined
\makeatother
\usepackage[colorlinks,citecolor=blue,urlcolor=blue,bookmarks=false,hypertexnames=true]{hyperref} 
\usepackage{natbib}
\usepackage{blindtext,tikz}
\usetikzlibrary{calc}

\algnewcommand\algorithmicforeach{\textbf{for each}}
\DeclareMathOperator*{\argmax}{arg\,max}

\algdef{S}[FOR]{ForEach}[1]{\algorithmicforeach\ #1\ \algorithmicdo}
\algnewcommand\algorithmicinput{\textbf{Input:}}
\algnewcommand\INPUT{\item[\algorithmicinput]}
\algnewcommand\algorithmicoutput{\textbf{Output:}}
\algnewcommand\OUTPUT{\item[\algorithmicoutput]}
\graphicspath{{./pics/}{./pics/field-exp/}{./pics/new-pics/}{./pics/pdf/}{./pics/simulation-exp/}}
\DeclareGraphicsExtensions{.pdf,.png,.jpg}

\usepackage{everypage}
\newcommand{\AddBackground}{
  \begin{tikzpicture}[remember picture,overlay]
    \node at ($(current page.north)+(0.0, -1.0)$) [rotate=0] {\textcolor{gray}{The manuscript is published in the International Journal of Robotics Research (IJRR) on January 19, 2024}};
  \end{tikzpicture}
}

\AddEverypageHook{\AddBackground}

\begin{document}
\title{Kernel-based Diffusion Approximated Markov Decision Processes for Autonomous Navigation and Control on Unstructured Terrains}
\author{Junhong Xu$^1$, Kai Yin$^2$, Zheng Chen$^1$, Jason M. Gregory$^3$, Ethan A. Stump$^3$, Lantao Liu$^1$
\thanks{$^1$Luddy School of Informatics, Computing, and Engineering  at Indiana University, Bloomington, IN 47408, USA. E-mail:
        {\tt\small \{xu14, zc11, lantao\}@iu.edu}.}
\thanks{$^2$Expedia Group. E-mail:
        {\tt\small kyin@expediagroup.com}.} 
\thanks{$^3$Army Research Laboratory. E-mail:
        {\tt\small \{jason.m.gregory1, ethan.a.stump2\}.civ@army.mil}.}
}
\maketitle
\IEEEpeerreviewmaketitle

\begin{abstract}
We propose a diffusion approximation method to the continuous-state Markov Decision Processes (MDPs) that can be utilized to address autonomous navigation and control in unstructured off-road environments. In contrast to most decision-theoretic planning frameworks that assume fully known state transition models, we design a method that eliminates such a  strong assumption that is often extremely difficult to engineer in reality.  We first take the second-order Taylor expansion of the value function. The Bellman optimality equation is then approximated by a partial differential equation, which only relies on the first and second moments of the transition model. By combining the kernel representation of the value function, we design an efficient policy iteration algorithm whose policy evaluation step can be represented as a linear system of equations characterized by a finite set of supporting states. We first validate the proposed method through extensive simulations in $2D$ obstacle avoidance and $2.5D$ terrain navigation problems. The results show that the proposed approach leads to a much superior performance over several baselines. We then develop a system that integrates our decision-making framework with onboard perception and conduct real-world experiments in both cluttered indoor and unstructured outdoor environments. The results from the physical systems further demonstrate the applicability of our method in challenging real-world environments.
\end{abstract}

\section{Introduction}

\begin{figure}
    \centering
    \includegraphics[width=0.98\linewidth]{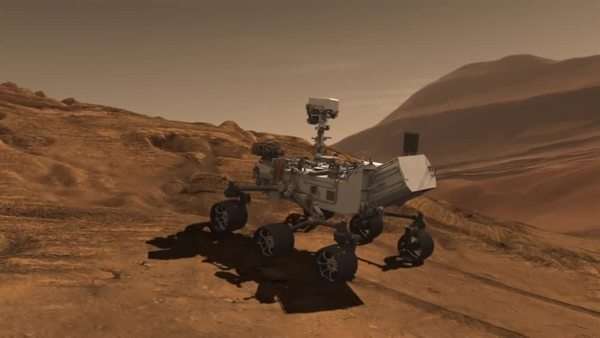}
    \caption{ In unstructured environments, the robot needs to make motion decisions in the navigable space with spatially varying terrestrial characteristics (hills, ridges, valleys, slopes). 
    This is different from the simplified and structured environments where there are only two types of representations, i.e., either obstacle-occupied or obstacle-free. 
    Evenly tessellating the complex terrain to create a discretized state space cannot effectively characterize the underlying value function used for computing the MDP solution. 
    (Picture credit: NASA)}
    \label{fig:mars} 
\end{figure}

The decision-making of an autonomous mobile robot moving in unstructured environments typically requires the robot to account for uncertain action (motion) outcomes and at the same time, maximize the long-term return.
The Markov Decision Process (MDP) is a very useful framework for formulating such decision-theoretic planning problems~\citep{boutilier1999decision}. 
Since the robot is moving in a continuous space, directly employing the standard form of MDP needs a discretized representation of the robot's state and action. 
For example, in practice, the discretized robot states are associated with spatial tessellation~\citep{thrun2000probabilistic}, and a grid-map-like representation has been widely used for robot planning problems where each grid is regarded as a discrete state; similarly, actions are simplified as transitions to traversable grids which are usually the  very few numbers of adjacent grids in the vicinity.

However, discretization  can be problematic. 
Specifically, if the discretization is low in resolution (i.e., large but few numbers of grids), the decision policy becomes a very rough approximation of the  simplified (discretized) version of the original problem;
on the other hand, if the discretization is high in resolution, the result might be approximated well, but this will induce prohibitive computational costs and prevent real-time decision-making. 
Finally, the characteristics of state space might be complex, and it is inappropriate to conduct lattice-like tessellation, which is likely to result in sub-optimal solutions. See Fig.~\ref{fig:mars} for an illustration.

Another critical issue lies in MDP's transition model, 
which describes the probabilistic transitions between states.  
However, obtaining an accurate probability distribution function for robot motion transitions is oftentimes unrealistic, even without considering spatiotemporal variability. 
This is another important factor that significantly limits the applicability of MDP in many real-world problems.  
Reinforcement learning~\citep{kober2013reinforcement} does not rely on a transition model specification but requires 
a large number of
training trials to learn the value function and the policy, which can be viewed as another strong and difficult 
assumption in many robotic missions. 
Thus, it is desirable that the demanding assumption of the known transition probability function 
can be relaxed as some non-exact form and can be obtained without cumbersome learning trials. 
This can be achieved by leveraging the major characteristics of the transition probabilistic distribution (e.g., moments or quantile of the distribution), 
which 
typically can be obtained (well approximated) from lightweight historical data or offline tests~\citep{thrun2000probabilistic}. 
If we only assume such ``partial knowledge" -- first two moments -- of the transition model, we must 
re-design the modeling and solving mechanisms. 

To address the above problems, we propose a diffusion-approximated and kernel-based solution. 
Our primary focus is to develop a theoretically-sound and practically-efficient solution to compute the optimal value function in continuous-state MDPs without requiring a full probabilistic transition model.

Our contributions can be summarized as follows:

\begin{itemize}
    \item 
    First, we apply the second-order Taylor expansion to the value function to relax the requirement of fully known transition functions. 
    The Bellman-type policy evaluation equation and the Bellman optimality equation are then approximated by a diffusion-type partial differential equation (PDE), which only relies on the first and second moments of the transition probability distribution.
    {The solution of the Taylored Bellman optimality equation can be viewed as an infinite-horizon discounted reward diffusion process. 
    }
    
    \item Second, 
    to improve both the efficiency and {the flexibility} of the value function {approximation}, 
    we use kernel functions which can represent a large number of function families for better value approximation.
    This approximation can conveniently characterize the underlying value functions with a finite set of discrete supporting states, leading to fast computation. 

    \item 
    Finally, we develop an efficient policy iteration algorithm by integrating the kernel value function representation and the Taylor-based diffusion-type approximation to Bellman optimality equation. 
    The policy evaluation step can be represented as a linear system of equations with characterizing values at the finite supporting states, and the only information needed is the first and second moments of the transition function.
    This alleviates the need for heavily searching in continuous/large state space and the need for carefully modeling/engineering the transition probability.
\end{itemize}

The organization of the paper is as follows:
we review the relevant literature and provide background in formulating stochastic planning problems and value function approximation methods in Section~\ref{sec:related-work} and~\ref{sec:preliminary}, respectively.
In Section~\ref{sec:method}, we present the Taylored Bellman equation and the proposed policy iteration algorithm.
Section~\ref{Experiments} provides evaluation results for various important algorithmic properties of the proposed method. 
Then, we integrate the perception into our framework to build an autonomy system and demonstrate real-world experiments in Section~\ref{sec:system}--\ref{sec:real-world-exp}. 
Finally, we conclude the paper and discuss insights learned from the real-world experiments in Section~\ref{sec:conclusion}. 
This paper extends our prior conference publication~\citep{Liu-RSS-20}  by providing technical details to justify the methods within Section~\ref{sec:method}, integrating up-to-date literature, and conducting extensive experiments, particularly in real-world settings delineated in Sections~\ref{sec:system}--\ref{sec:real-world-exp}.


\section{Related Work}\label{sec:related-work}
Our work closely relates to value-based methods for solving Markov Decision Processes (MDPs) and stochastic path and motion planning in robotics.
In this section, we provide concise surveys of the two fields respectively and highlight our contributions at last. 

\subsection{Value-Based Methods for Solving Continuous-State MDPs}
Motion planning under action uncertainty can be framed as an MDP with continuous state and action~\citep{bertsekas2012dynamic, sutton2018reinforcement}.
Many existing works rely on value-based methods for solving MDPs. 
This approach focuses on computing a value function from which the policy is derived implicitly using the Bellman optimality equation. 
However, the intractability of an exact value function representation emerges when dealing with a large or continuous state space, as it requires enumerating over an infinite-sized table that stores every state value.
Thus, a large body of research focuses on value function approximation techniques~\citep{kober2013reinforcement, sutton2018reinforcement}.

One simple approach to approximating the value function is tessellating the continuous state space into finite uniform grids.
This method is popular in many robotic applications~\citep{pereira2013risk, otte2016any, fu2015sense, al2013wind, baek2013optimal}. 
However, this uniform tessellation approach does not scale well to environments with complex geometric structures, and it requires a large number of grids when the problem size increases, known as the curse of dimensionaliy~\citep{bellman2015adaptive}.
A more advanced discretization technique that alleviates this problem is by adaptive discretization~\citep{gorodetsky2015efficient, liu2018solution, Liu-RSS-19, munos2002variable, lavalle2006planning}.

Alternative methods tackle this challenge by representing and also approximating the value function by a linear combination of basis functions or some parametric functions~\citep{bertsekas1996neuro, sutton2018reinforcement, munos2008finite}. 
The weights in the linear combination can be optimized by minimizing the Bellman residual~\citep{antos2008learning}.
However, these methods do not apply to complicated problems because selecting a proper set of basis functions is non-trivial.
This weakness may be resolved through kernel methods~\citep{hofmann2008kernel}.
Because the weights in a linear combination of basis functions can be presented by their product (through the so-called duel form of least squares~\citep{shawe2004kernel}), the product may be better replaced by kernel functions (on so-called reproducing kernel Hilbert spaces) at supporting states. 
Once value functions at supporting states are obtained, the approximation to the value function at any state is also determined. 
In~\citet{deisenroth2009gaussian}, the authors use a Gaussian process with a Radial Basis Kernel (RBF) to approximate the value function.
This approach to approximating the value function by kernel functions is referred as the {\em direct kernel-based method} in this paper. 
There is a vast literature on kernelized value function approximations in reinforcement learning~\citep{engel2003bayes, kuss2004gaussian, taylor2009kernelized, xu2007kernel}.
But few studies in robotic planning problems leveraged this approach. 
A recent application of work~\citep{engel2003bayes} on marine robots can be found in~\cite{martin2018sparse}.

More recently, due to the advancement of deep learning, neural networks are proposed to  approximate value functions in high-dimensional state spaces~\citep{arulkumaran2017brief} such as images~\citep{nair2018visual} and high-dimensional control problems~\citep{lillicrap2015continuous}.
Additionally, many state-of-the-art reinforcement learning (RL) algorithms integrate value-based and policy gradient methods, known as actor-critic~\citep{schulman2017proximal}.
Unlike the value-based method, whose policy is derived implicitly, the actor-critic framework learns a policy directly using the policy gradient computed from the value function approximated by a neural network.
Through the integration of physics simulation, the techniques within this framework have demonstrated successful real-world applications, particularly in controlling locomotion across unstructured terrains~\citep{makoviychuk2021isaac,wang2021rough, miki2022learning,das2022simulation}. 
These methods rely on sampling to compute the value function update via the Bellman equation.
It can be sample-inefficient because numerous next-state samples are typically required to compute a Bellman update.
In contrast, our proposed method increases the sample efficiency by converting the Bellman equation to a PDE.
This operation eliminates the need for sampling next states, requiring only the evaluation of partial derivatives for Bellman updates, bypassing the potentially expensive sampling process. 
We demonstrate this property in Section~\ref{Experiments}.

Furthermore, all the above-mentioned existing approaches rely on fully known transitions in MDP, require samples from the complex stochastic motion model, or rely on careful selection of basis functions, which are difficult to design in practice.
Therefore, the challenge becomes {\em how to design a principled methodology without explicitly relying on basis functions and without full knowledge of transitions in MDP, which will be addressed in this work}.

\subsection{Deterministic and Stochastic Planning in Robotics}\label{literature-planning}
Most of the traditional robotic planning methods (see~\citet{gammell4asymptotically} for a recent survey) use a simplified deterministic robot motion model, e.g., holonomic kinematics, to compute an open-loop path, e.g., sampling-based motion planners such as probabilistic roadmap (PRM) and rapidly exploring random tree (RRT) and their variants~\citep{lavalle2006planning, karaman2011sampling}. 
Since the modeling error exists between the model used in planning and the real world, the robot cannot execute the path directly and requires a separate controller to track the path. 
To ensure the path is trackable by the controller, refining the planned path into a kinodynamically trackable trajectory~\citep{webb2012kinodynamic} is necessary.
{One strategy to generate such trajectories is based on trajectory optimization.} 
Existing trajectory optimization methods can be categorized into hard-constrained methods and soft-constrained methods. Using hard-constrained optimization technique to solve trajectory generation problem has been proposed by \citet{mellinger2011minimum}, where piecewise polynomial trajectories are generated by solving a quadratic programming (QP) problem. A closed-form solution is provided to solve an unconstrained QP problem, and intermediate waypoints are added iteratively to ensure the safety of the trajectories \citep{richter2016polynomial}. Free space represented by multiple geometric volumes, e.g., cubes \citep{chen2015real, gao2018online}, spheres \citep{gao2016online, gao2019flying}, and polyhedra \citep{liu2017planning, deits2015efficient}, are used to formulate a convex optimization problem, which generates smooth trajectories within the volumes. However, hard-constrained methods ignore the distance information to the obstacle boundaries and are prone to generate trajectories close to the obstacles. This can cause collisions when the robot motion model is imperfect. Such safety issue motivates the development of the soft-constrained methods which leverage the distance gradient during optimization such that the trajectories could be adjusted to stay far from obstacle boundaries while maintaining the smoothness property \citep{zhou2020robust, oleynikova2016continuous, zhou2019robust}.

The above design of the navigation system is based on deterministic models at all levels, i.e., from path planning to trajectory optimization, and uncertainties in the motion are entirely handled by the tracking controller. 
Such decoupled design is brittle because it is possible that the feedback controller cannot stabilize on the open-loop path computed by the planner.
Therefore, to reduce the burden of designing and tuning separate controllers for different environments, it is desirable to develop methods that directly consider the action execution uncertainty during the planning phase if the error between the planning model and the real-world action is large. 
This problem is particularly obvious when the robot moves on rough terrains.

Early work of stochastic planning in the robotics community was mostly built upon the sampling-based motion planning paradigm.
In~\citet{tedrake2010lqr}, the authors compute local feedback controllers while adding new tree branches in the RRT algorithm to stabilize the robot's motion between two tree nodes. 
A similar idea has also been exploited by~\citet{agha2014firm} with a different planner and controller.
This method considers both the sensing and motion uncertainty and uses the linear-quadratic Gaussian (LQG) controller~\citep{kalman1960contributions} to stabilize the motion and sensing uncertainties along the edges generated by a PRM algorithm.
Alternatively, \citet{van2011lqg} use the LQG controller to compute a distribution of paths,  from which the path is planned by the RRT algorithm that optimizes an objective function based on the path distribution. 
Instead of modular design of the planning and control components, \citet{huynh2016incremental} propose a more integrated approach to solving the general continuous-time stochastic optimal control problem.
The method first approximates the original MDP by a discrete MDP through sampling points in the state space, and then a closed-loop policy is computed in this discrete MDP. 
The approximation and solution are then refined by iterating this sampling and solving procedure. 

Model predictive control (MPC), or receding horizon control, is another strategy to resolve motion uncertainty~\citep{rawlings2017model, bertsekas2005dynamic}. 
At each timestep, it solves a finite horizon optimal control problem and applies the first action of the computed open-loop 
trajectory online. 
To reduce computational burden, MPC usually uses a deterministic model to predict future states~\citep{bertsekas2012dynamic}.
Robust MPC~\citep{bemporad1999robust} explicitly deals with the model uncertainty by optimizing over feedback policies rather than open-loop trajectories.
However, this optimization is computationally expensive.
Thus, in practice, tube MPC that focuses only on a sub-space is often used as an approximate solution to robust MPC~\citep{langson2004robust}. 
Tube MPC for nonlinear robot motion models is still an active research area.
For example, sum-of-squared programming~\citep{majumdar2013robust, majumdar2017funnel} and reachibility analysis~\citep{althoff2008reachability, bansal2017hamilton} are used for solving the nonlinear Tube MPC problem. 

In general, robot motion planning essentially requires to reason about uncertainties during motion control.
However, because the prohibitive computation prevents the robot from real-time decision, most of the traditional approaches to generating feasible trajectories do not consider uncertainties.
Oftentimes some heuristics are leveraged to ensure the motion safety during execution, e.g., using the soft-constrained methods. 
In contrast, our proposed framework solves the stochastic planning problem in a direct and integrative fashion. 
As a result, the computed policy can naturally guide the robot with a large clearance from the obstacles,  without adding extra penalty in the objective function which can be hard to specify in practice. 
In addition, most existing methods that solve the stochastic planning problems 
avoid the difficulties in directly computing the optimal value function through using arbitrary, {e.g., non-linear or discontinuous}, reward and transition functions. 
Different from that, the proposed approach makes the direct computation possible in practice, which opens a new line of work to solve stochastic motion planning problems.

Emerged as a popular sampling-based MPC approach, 
Model Predictive Path Integral (MPPI) control~\citep{kappen2012optimal, theodorou2010generalized, williams2017information, theodorou2010generalized, williams2018robust} also exhibits relevance to our proposed method. 
This methodology has been developed specifically to address finite-horizon stochastic optimal control problems that conveniently allow arbitrary cost functions and nonlinear system dynamics. It shares a fundamental objective with our method, focusing on the resolution of the second-order Hamilton–Jacobi–Bellman (HJB) partial differential equation. Notably, MPPI correlates between noise and control costs, utilizing the Feynman–Kac formula to transform the HJB equation into a path integral formulation. This transformation enables MPPI to provide an MPC solution via a ``forward" sampling strategy.
Contrastingly, our method diverges in approach by leveraging a kernel representation of the value function. This representation forms the foundation for our methodology to directly compute solutions to the HJB equation via a policy iteration process. 
This differs our method from the MPPI framework and lays the groundwork for potential advancements and optimizations within the realm of stochastic optimal control problems.


\section{Preliminary Background}\label{sec:preliminary}

\subsection{Markov Decision Processes}
We formulate the robot decision-theoretic planning problem as an infinite horizon discrete time discounted Markov Decision Process (MDP) with continuous states and finite actions. 
It is defined by a 5-tuple $\mathcal{M}\triangleq\langle\mathbb{S}, \mathbb{A}, T, R, \gamma\rangle$,
where $\mathbb{S}=\{s\} \subseteq \mathbb{R}^{d}$ is the $d$-dimensional continuous state space and
$\mathbb{A} = \{a\}$ is a finite set of actions. $\mathbb{S}$ can be thought of as the robot workspace in our study. 
A robot transits from a state $s$ to the next $s'$ by taking an action $a$ in a stochastic environment and obtains a reward $R(s, a)$. 
Such transition is governed by a conditional probability distribution $T(s, a, s') \triangleq p(s'|s, a)$ which is termed as transition model (or transition function); the reward $R(s, a)$, a mapping from a pair of state and action to a scalar value, which specifies the one-step objective that the robot receives by taking action $a$ at state $s$.
The final element $\gamma \in (0, 1)$ in $\mathcal{M}$ is a discount factor which will be used in the expression of value function.

We consider the class of deterministic policies $\Pi$, which defines a mapping $\pi \in \Pi: \mathbb{S}\rightarrow \mathbb{A}$ from a state to an action. 
The expected discounted cumulative reward for any policy $\pi$ starting at any state $s$ is expressed as
\begin{align}\label{eq:value-fn-policy}
    v^{\pi}(s) = \mathbb{E}\left[\sum_{t=0}^{ \infty} \gamma^{t} R(s_t, \pi(s_t))|s_0 = s\right]. 
\end{align}
The state at the next time step, $s_{t+1}$,  draws from distribution $p(s_{t+1} | s_t, \pi(s_t))$. 
Let us use $k$ to denote the computation epoch, 
then we can rewrite the above equation recursively as follows
\begin{equation}\label{eq:value-function}
\begin{split}
    v_{k+1}^{\pi}(s) =& R(s, \pi(s)) + \gamma\,\mathbb{E}^{\pi}[v_k^{\pi}(s')|s] \triangleq \mathcal{B}^{\pi}v_k^{\pi}(s),
\end{split}
\end{equation}
where 
$\mathcal{B}^{\pi}$ is called the {\em Bellman operator}, and $\mathbb{E}^{\pi}[v_k^{\pi}(s')|s]=\int p(s'|s, \pi(s))v_k^{\pi}(s')\,ds'$. 
The computation epoch $k$ is omitted in the rest of the paper for notation simplicity. 
Eq.~\eqref{eq:value-function} is called Bellman equation.
The function $v^{\pi}(s)$ is usually called the {\em state value function} of the policy $\pi$. 
Solving an MDP amounts to finding the optimal policy $\pi^*$ with the optimal value function which satisfies the \textit{Bellman optimality equation} 
\begin{equation}\label{bellman-equation}
  v^{\pi^*}(s) = \max_{\pi}\left\{R(s, \pi(s)) + \gamma\,\mathbb{E}^{\pi}[v^{\pi}(s')|s]\right\}.
\end{equation}

\subsection{Approximate Policy Iteration via Value Function Representation}\label{sec:primal}

Value iteration and policy iteration are the most prevalent approaches to solving an MDP.
It has been shown that the value iteration and policy iteration can both converge to the optimal policy in MDPs with discrete states~\citep{bertsekas2012dynamic, sutton2018reinforcement}. 
Our work will be built upon policy iteration and here 
we provide a summary of the common value function approximation methods used in policy iteration when dealing with continuous states~\citep{powell2016perspectives,gordon1999approximate, bertsekas2011approximate}. 

Policy iteration requires an initialization of the policy (can be random).
When the number of states in the MDP is finite, a system of finite number of linear equations can be established based on the initial policy, where each equation is exactly the value function (Eq.~\eqref{eq:value-function}). 
The solution to this linear system yields state-values for all states of the initial policy~\citep{puterman2014markov}. 
This step is called \textit{policy evaluation}. The second step is to improve the current policy by greedily improving local actions based on the incumbent values obtained. This step is called \textit{policy improvement}. 
Through iterating these two steps, we can find the optimal policy and a unique solution to the value function that satisfies Eq.~\eqref{eq:value-fn-policy} for every state. 

If, however, the states are continuous or the number of states is infinite, it is intractable to store and evaluate 
the value function at every state. 
One must resort to approximate solutions by finding an appropriate representation of the value function. 
Suppose that the value function can be represented by a weighted linear combination of known functions where only weights are to be determined, 
then a natural way to go is leveraging the Bellman-type equation, i.e., Eq.~\eqref{eq:value-function}, to compute the weights. 
Specifically, given an arbitrary policy, the representation of value function can be evaluated at a finite number of states, leading to a linear system of equations whose solutions can be viewed as weights~\citep{lagoudakis2003least}. 
This obtained representation of the value function can be used to improve the current policy. The remaining procedure is then similar to the standard policy iteration method.
The final obtained value function representation serves as an approximated optimal value function \textit{for the whole continuous state space}, and the corresponding policy can be obtained accordingly. 

Formally, let the value function approximation under policy $\pi$ be 
\begin{equation}\label{basis-approx}
 v^{\pi}(s)\simeq v(s; w^{\pi}) = \sum_{i=1}^m w_i^{\pi}\cdot \phi_i(s),    
\end{equation}
where $\phi_i\in\Phi\triangleq\{\phi_1, \ldots, \phi_m\}$. 
The elements in the set $\Phi$ are called the \textit{basis functions} in literature~\citep{powell2016perspectives}, and these basis functions are usually parametric functions with a fixed form. 
A finite number of supporting states $\mathbf{s} = \{s^{1}, \ldots, s^N\}$, $N>m$ can be selected.
The selection of supporting states $\mathbf{s}$ needs to take into account the characteristics of the underlying value functions.
In the scenario of robot motion planning in complex terrains, it relates to the properties of the terrain, e.g., geometry or texture. 
The solution to  $w^{\pi}$  can be calculated by minimizing the 
squared {\em Bellman error} over $\mathbf{s}$, defined by 
$\mathcal{L}(w^{\pi}) = \sum_{i=1}^{N}({v(s^i;w^{\pi}) - \mathcal{B}^{\pi}v({s}^i; w^{\pi}))^2}.$
And $w^{\pi}$ may have a closed form solution in terms of the basis functions, transition probabilities, and rewards~\citep{lagoudakis2003least}. By policy iteration, the final solution for $v(s; w^{\pi})$ can be obtained. Note that $v(s)$ may also be  approximated by any non-parametric nonlinear functions such as neural networks.

\section{Methodology}\label{sec:method}
Our objective is to design a principled kernel-based policy iteration approach by leveraging kernel methods to solve the continuous-state MDP. 
In contrast to most decision-theoretic planning frameworks which assume fully known MDP transition probabilities~\citep{boutilier1999decision, puterman2014markov}, 
we propose a method that eliminates such a strong premise which oftentimes is  extremely difficult to engineer in practice. 
To overcome this challenge, 
first we apply the second-order Taylor expansion of the kernelized value function (Section~\ref{sec:taylored-policy-eval}). 
The Bellman optimality equation is then approximated by a partial differential equation which only relies on the first and second moments of transition probabilities (Section~\ref{sec:bellman-eqn-pde}). 
Combining the kernel representation of value function, this approach efficiently tackles the continuous or large-scale state space search with  minimum prerequisite knowledge of state transition model (Sections~\ref{kernel-taylored-P-E} and ~\ref{K-T-P-I}).

\subsection{Taylored Approximate Bellman Equation}
\label{sec:taylored-policy-eval}

To design an efficient approach for solving continuous-state MDP problems, we essentially need to fulfill two requirements: {\it a suitable representation of the value function} and {\it efficient computation of the Bellman optimality equation}. 

For the first requirement, we may directly apply the basis functions to approximate the value function and then solve the Bellman optimality equation. Yet this approach faces difficulties of explicitly specifying basis functions: if the set of basis functions is not rich enough, the approximation error can be large. A better approach may be a direct application of kernel methods to represent the value function (referred as the {\em direct kernel-based method}). We will discuss this method later.   
But this representation does not satisfy the second requirement, i.e., {\em efficient evaluation of the Bellman optimality equation}.
This is because a fully specified transition probability function $p(\cdot | \cdot, \cdot)$ is required for computing the Bellman operator, but it is usually hard to specify in the real world.
In addition, even this transition function can be obtained, the expectation $\mathbb{E}^{\pi}[v^{\pi}_{k}(s')] = \int p(s'|s,\pi(s)) v_k^{\pi}(s') ds'$ generally does not have a closed-form solution, and computationally expensive numerical integration is typically entailed.

In contrast to directly solving Eq.~(\ref{bellman-equation}) by value function approximation, we consider an approximation to the Bellman-type equation at first and then apply value function approximation. 
We approximate the Bellman equation by using only first and second moments of transition functions.
This will allow us to obtain a nice property that a complete and accurate transition model is not necessary; instead, only the important statistics such as mean and variance (or covariance) will be sufficient for most real-world applications.
Additionally, we will see that evaluating the integration over the global state space is not needed. From this perspective, our approximation can be viewed as a local version of the original Bellman equation, 
{i.e., it uses local gradient information to approximate the integral of the value function over possible next states}. 

Formally, we assume that the state space is of $d$-dimension and a state $s$ may be expressed as $s=[s_1, s_2, \ldots, s_d]^T$.
Suppose that the value function $v^{\pi}(s)$ for any given policy $\pi$ has continuous first and second order derivatives (this can usually be satisfied with aforementioned value function designs). 
We subtract both hand-sides by $v^{\pi}(s)$ from Eq.~\eqref{eq:value-function} and 
then take Taylor expansions of value function around $s$ up to second order~\citep{braverman2018taylor}: 
\begin{align}
    &-R(s, \pi(s)) \nonumber\\
    &=\gamma\left(\mathbb{E}^{\pi}[v^{\pi}(s')\mid s] - v^{\pi}(s)\right) - (1-\gamma)\,v^{\pi}(s) \nonumber\\
    &=\gamma\int p(s'| s,\pi(s))(v^{\pi}(s') - v^{\pi}(s))\,ds' - (1-\gamma)\,v^{\pi}(s) \nonumber\\
    &\simeq \gamma\,\Big((\mu^{\pi}_s)^T\,\nabla v^{\pi}(s) +\frac{1}{2}\nabla\cdot \sigma_s^{\pi}\nabla v^{\pi}(s)\Big) - (1-\gamma)\,v^{\pi}(s), \label{bellman-type-pde}
\end{align}
where $\mu^{\pi}_s$ and $\sigma^{\pi}_s$ are the first moment (i.e., a $d$-dimensional vector) and the second moment (i.e., a $d$-by-$d$ matrix) of transition functions, respectively, with the following form
\begin{subequations}\label{mu-sigma-eqns}
\begin{align}
    (\mu^{\pi}_s)_i &= \int p(s'| s,\pi(s))(s'_i - s_i)\,ds', \\
    (\sigma^{\pi}_s)_{i, j} &= \int p(s' | s,\pi(s))(s'_i - s_i)(s'_j - s_j)\,ds',
\end{align}
\end{subequations}
for $i, j\in\{1, \ldots, d\}$; 
the operator $\nabla \overset{\Delta}{=} [\partial/\partial s_1, ..., \partial/\partial s_d]^T$ and the notation $\cdot$ in the last equation indicate an inner product; the operator $\nabla\cdot \sigma^{\pi}_s\nabla$ is read as
\begin{eqnarray}\label{sigma_expr}
    \nabla\cdot \sigma^{\pi}_s\nabla = \sum_{i,j}(\sigma^{\pi}_{s})_{i,j}\frac{\partial^2}{\partial s_i\partial s_j}. 
\end{eqnarray}

To be concrete, we take a surface-like terrain for example and use that surface as the decision-theoretic planning workspace, i.e., $s=[x,y]^T\,\overset{\Delta}{=}\,[s_x, s_y]^T$. We have the expression for the following operator 
\begin{eqnarray*}
    \nabla\cdot \sigma^{\pi}_s\nabla = \sigma^{\pi}_{xx}\frac{\partial^2}{\partial x^2} +\sigma^{\pi}_{xy}\frac{\partial^2}{\partial x\partial y}
    +\sigma^{\pi}_{yx}\frac{\partial^2}{\partial y\partial x}
    +\sigma^{\pi}_{yy}\frac{\partial^2}{\partial y^2}.
\end{eqnarray*}

Since Eq. (\ref{bellman-type-pde}) approximates calculation of Eq.~\eqref{eq:value-function} in the policy evaluation stage, the solution to Eq. (\ref{bellman-type-pde}) thus provides the value function approximation under current policy $\pi$. 
Eq.~\eqref{bellman-type-pde} also implies that we only need to use the first $(\mu^{\pi}_s)_i$ and second $(\sigma^{\pi}_s)_{i, j}$ moments instead of computing the expectation of the value function using the original transition model $p(s'| s,\pi(s))$ to evaluate the Bellman operator Eq.~\eqref{bellman-equation} required in solving MDPs.

Note that our method can be naturally extended to include higher moments via higher-order Taylor expansions.  
We stop at the second moment because the first two moments are generally sufficient to describe the major characteristics of the transition function. Also, extending to higher moments will incur heavy computation which is a trade-off that we need to take into account.

\subsection{Approximate Bellman Optimality Equation via Diffusion-type PDE}\label{sec:bellman-eqn-pde}

Eq.~\eqref{bellman-type-pde} is a diffusion-type partial differential equation (PDE).
It is an approximation to the Bellman equation~\eqref{diffusion-pde} through the second-order Taylor expansion. We will need to solve this PDE to obtain 
an approximation methodology to the Bellman optimality equation.
Typically if solutions exit for a PDE, there could be infinite solutions to satisfy the 
the PDE unless we impose proper boundary conditions~\citep{Evens2010}. Therefore, we need to analyze the necessary boundary conditions to Eq.~\eqref{bellman-type-pde}.

\begin{figure}[t]
    \centering
    \includegraphics[width=0.9\linewidth]{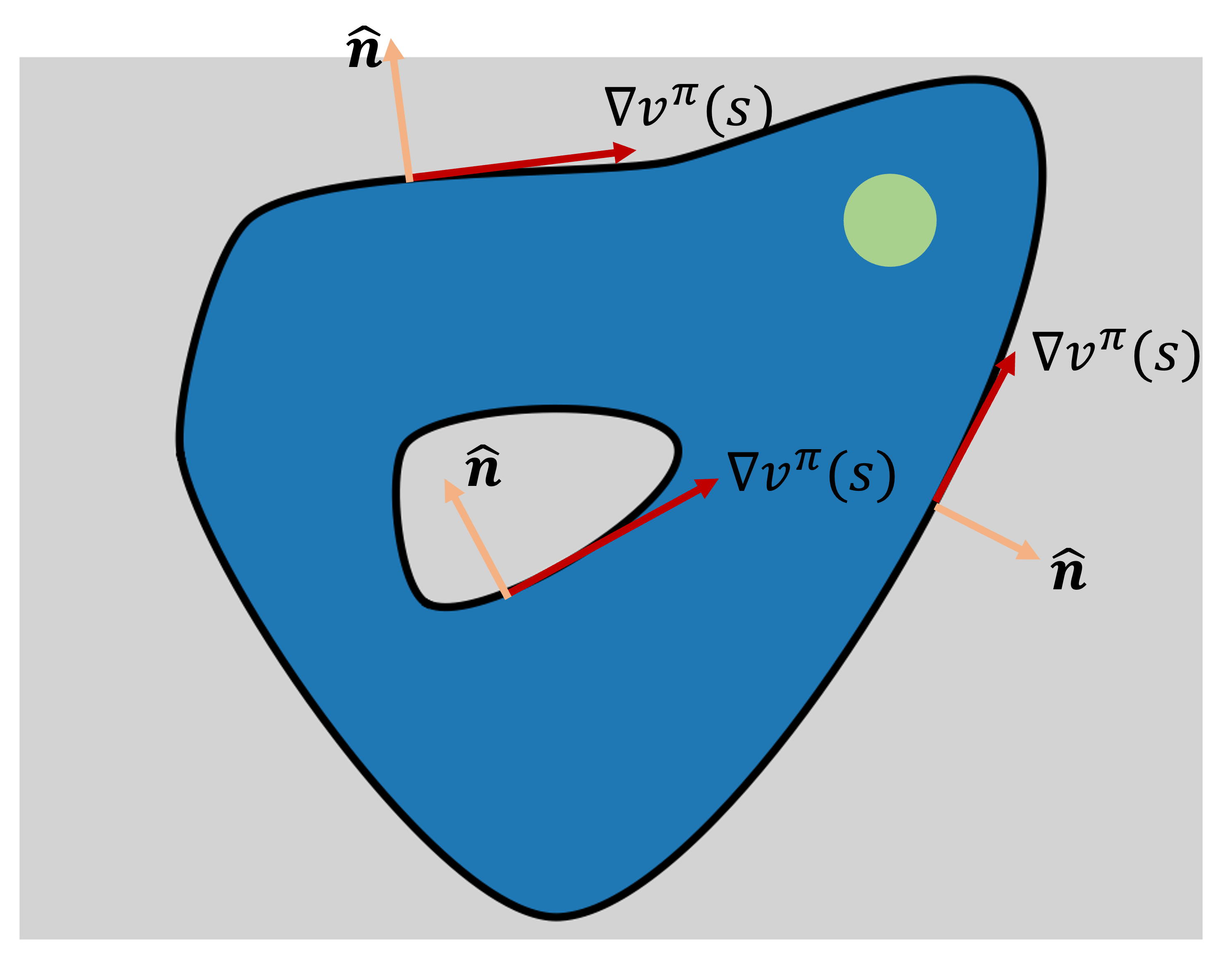}
    \caption{{Illustration of the boundary condition in Eq.~\eqref{boundary-condition-a} using a 2D example. 
    The blue and green regions indicate the state space and the goal region, respectively.
    The grey areas represent the infeasible space, e.g., obstacles, and the boundaries are shown as black curves.
    The normal vector $\hat{\mathbf{n}}$ and the gradient of value function $\nabla v^{\pi}(s)$ at three arbitrary boundary points are indicated by yellow and red arrows, respectively.}
    }
    \label{fig:boundary-illustration}
\end{figure}

In our problem settings, robots are not allowed to move out of free-space boundaries. We first observe that the value function should not have values outside of the feasible planning region, i.e., the state space $\mathbb{S}$, and the value function should not increase towards the boundary of the region. 
Otherwise it will result in actions that guide the robot outside the free space. A practical boundary condition to impose can be that the directional derivative of the value function with respect to the outward unit normal at the boundary states is zero (see Fig.~\ref{fig:boundary-illustration} for an illustration). It is not the only condition that we can impose, but it is a relatively easier condition to obtain solutions with desired behaviors. 
Second, in order to ensure that the robot is able to reach the goal, we follow the conventional goal-oriented decision-theoretical planning setup and constrain the value function at the goal state to the maximum state-value in the state space $\mathbb{S}$. 
Consequently, these boundary conditions ensure that the policy does not control the robot outside of the feasible regions (safety) and also leads the robot to the goal area. 

Formally, let us denote the boundary of entire continuous planning region 
by $\partial\mathbb{S}$ and the goal state by $s_g$. Suppose the value function at $s_g$ is $v_{g}$.
Section~\ref{sec:taylored-policy-eval} implies that the Bellman optimality equation Eq.~\eqref{bellman-equation} can be approximated by the following {Hamilton–
Jacobi–Bellman (HJB)} PDE:
\begin{eqnarray}
 0&=&\max_{\pi}\Big\{\gamma\,\Big((\mu_s^\pi)^T\nabla v^{\pi}(s) +\frac{1}{2}\nabla\cdot \sigma_s^\pi\nabla v^{\pi}(s)\Big) + \nonumber\\
 &{}& R(s, \pi(s)) - (1-\gamma)\,v^{\pi}(s)\Big\}, 
 \label{diffusion-pde}
\end{eqnarray}
with boundary conditions
\begin{subequations}\label{boundary-conditions}
\begin{eqnarray}
    \sigma_s^{\pi}\,\nabla v^{\pi}(s)\cdot \hat{\bm{n}} &=& 0, \mbox{  on } \partial\mathbb{S} \label{boundary-condition-a}\\
    v^{\pi}(s_g) &=& v_g,  \label{boundary-condition-b}
\end{eqnarray}
\end{subequations}
where $\hat{\bm{n}}$ denotes the unit vector normal to $\partial\mathbb{S}$ pointing outward, {and condition~\eqref{boundary-condition-a} constrains the directional derivative with respect to this normal vector $\hat{\bm{n}}$ to be zero.} 
{The solution $v^{\pi}$ of the above PDE can be interpreted as a diffusion process with $\mu_s^{\pi}$ and $\sigma_s^{\pi}$ as drift and diffusion coefficients, respectively.} 
{We show a 2D example of this boundary condition in Fig.~\ref{fig:boundary-illustration}.
The normal vectors $\hat{\bm{n}}$ at the boundary are perpendicular to the gradient of the value function $\nabla v^{\pi}(s)$, which constrains the value function from expanding outside the boundaries of the state space.
In addition, the direction of each gradient vector points toward the goal, and this allows the policy to follow the value function gradient which guides the robot to move in the goal direciton.} 

The condition (\ref{boundary-condition-a}) is a type of homogeneous Neumann condition, and condition (\ref{boundary-condition-b}) can be thought of as a Dirichlet condition in literature \citep{Evens2010}.
This elegantly approximates the classic Bellman optimality equation by a convenient PDE representation. While Eq.~(\ref{diffusion-pde}) is a nonlinear PDE (due to the maximum operator over all the policies), our algorithm in the future section will allow to solve a linear PDE for a fixed policy.
In the next section, we will leverage the kernelized representation of the value function to avoid the difficulties of directly solving PDE. The kernel method will help transform the problem to a linear system of equations with 
unknown values at the finite supporting states.

\subsection{Kernel Taylor-Based Approximate Policy Evaluation}\label{kernel-taylored-P-E}
With aforementioned formulations, another critical research question is whether the value function can be represented by some special functions that are able to approximate large function families in a convenient way.
We tackle this question by using a kernel method to represent the value function. 
Thanks to Eq.~\eqref{diffusion-pde} which allows us to extend with kernelized policy evaluation for Taylored value function approximation. 

Specifically, let $k(\cdot, \cdot)$ be a generic kernel function$^{\dagger}$~\citep{hofmann2008kernel}.
\footnotetext{$^{\dagger}$ It is worth mentioning that our approach of utilizing kernel methods is to approximate the function. This usage should be distinguished from that in the machine learning literature where kernel methods are used to learn patterns from data.}
For a set of selected finite supporting states $\mathbf{s}=\{s^{1}, \ldots, s^N\}$, let $\mathbf{K}$ be the Gram matrix with $[\mathbf{K}]_{i,j}=k(s^i, s^j)$, and $\mathbf{k}(\cdot, \mathbf{s})=[k(\cdot, s^1), \ldots, k(\cdot, s^N)]^T$. Given a policy $\pi$, assume the value functions at $\mathbf{s}$ are   $V^{\pi}=[v^{\pi}(s^1), \ldots, v^{\pi}(s^N)]^T$. Then, for any state $s'$, the kernelized value function has the following form
\begin{equation}\label{kernerlized-value-func}
    v^{\pi}(s') = \mathbf{k}(s', \mathbf{s})^T\,\left(\lambda\mathbf{I}+\mathbf{K}\right)^{-1}\,V^{\pi},
\end{equation}
where $\lambda \geq 0$ is a {\em regularization factor}. When $\lambda = 0$, it links to the kernel ordinary least squares estimation of $w^{\pi}$ in Eq.~\eqref{basis-approx}; when $\lambda > 0$, it refers to the ridge-type regularized kernel least squares estimation~\citep{shawe2004kernel}. 
Furthermore, Eq.~\eqref{kernerlized-value-func} implies that as long as the values $V^{\pi}$ are available, the value function for any state can be immediately obtained. Now our objective is to get $V^{\pi}$ through Eq.~\eqref{bellman-type-pde} and boundary conditions Eq.~\eqref{boundary-conditions}.

Plugging the kernelized value function representation into Eq.~\eqref{bellman-type-pde}, we end up with the following linear system:
\begin{equation}\label{linear-system-V}
\left(\mathbf{M}^{\pi} \left(\lambda\mathbf{I}+\mathbf{K}\right)^{-1} -(1-\gamma)\,\mathbf{I}\right)\,V^{\pi} = \mathbf{R}^{\pi},
\end{equation}
where $\mathbf{I}$ is an identity matrix, $\mathbf{R}^{\pi}$ is a $N\times 1$ vector with element $[\mathbf{R}^{\pi}]_i = -R(s^i, \pi(s^i))$, and $\mathbf{M}^{\pi}$ is a matrix whose elements are:
\begin{equation}\label{equations-for-M}
[\mathbf{M}^{\pi}]_{i,j} = \gamma\left ((\mu^{\pi}_{s^i})^T\nabla_{s^i} + \frac{1}{2}\nabla_{s^i}\cdot \sigma_{s^i}^{\pi}\nabla_{s^i} \right)k(s^i, s^j).
\end{equation}
Note that $\nabla_{s^i}$ indicates the derivatives with respect to $s^i$, i.e., $\nabla_{s^i}\overset{\Delta}{=} [\partial/\partial s^i_1, \ldots, \partial/\partial s^i_d]^T$. 
Here, we provide a concrete derivation of using the Gaussian kernel to our proposed kernel Taylor-Based approximate method
as it is a commonly used kernel in practice and often used in the studies of kernel methods. 

Gaussian kernel functions on states $s'$ and $s$ have the form $k(s', s) = c\times\exp\left((-\frac{1}{2}(s'-s)^T\Sigma^{-1}_s(s'-s)\right)$, where $c$ is a constant and $\Sigma_s$ is a covariance matrix. Note that $\Sigma_s$ is referred to as the length-scale parameter in our work. The lengthscale governs the ``smoothness" of the function, and a large lengthscale leads to a smooth function whereas a small lengthscale causes a rugged function. 
Due to limited space, we only provide formula below for the first and second derivatives of the Gaussian kernel functions. These formula are necessary when Gaussian kernels are employed (for example, Eq.\eqref{equations-for-M}). In fact, we have 
\begin{equation}\label{Gaussian-first-dev}
    \nabla_{s'}k(s', s) = -\Sigma_s^{-1}(s'-s)k(s', s),
\end{equation}
and
\begin{equation}\label{Gaussain-sec-dev}
\begin{split}
   & \nabla_{s'}\cdot \sigma_s\nabla_{s'} k(s', s) = - tr(\sigma_s\,\Sigma_s^{-1})\,k(s', s)\\ 
   &+(s'-s)^T\Sigma_s^{-T}\sigma_s\,\Sigma_s^{-1}(s'-s)k(s', s),
\end{split}
\end{equation}
where $tr(\cdot)$ denotes the trace of the matrix.
By plugging Eq.~\eqref{Gaussian-first-dev} and Eq.~\eqref{Gaussain-sec-dev} into Eq.~\eqref{equations-for-M}, we can obtain an expression for the elements in the matrix $\mathbf{M}^{\pi}$. 

The solutions to the system Eq.~\eqref{linear-system-V} yield values of $V^{\pi}$. 
These values further allow us to obtain the value function~\eqref{kernerlized-value-func} for any state under current policy $\pi$. 
This completes modeling our kernel Taylor-based approximate policy evaluation framework.

\subsection{Kernel Taylor-Based Approximate Policy Iteration}\label{K-T-P-I}

\begin{algorithm}[htbp!] 
\caption{Kernel Taylor-Based Approximate Policy Iteration}
\label{alg:kernel-based-policy-iteration}
\begin{algorithmic}[1]
    \INPUT{
    A set of supporting states $\mathbf{s} = \{\mathbf{s}^1, ..., \mathbf{s}^N\}$;
    the kernel function $k(\cdot, \cdot)$; the regularization factor $\lambda$; 
    the MDP $\langle \mathbb{S}, \mathbb{A}, T, R, \gamma \rangle$. 
    }
    \OUTPUT{The kernelized value function Eq.~\eqref{kernerlized-value-func} for every state and corresponding policy.}
    \State Initialize the action at the supporting states.
    \State Compute the matrix $\mathbf{K} + \lambda \mathbf{I}$ and its inverse. 
    \Repeat
        \State // Policy evaluation step
        \State Solve for $V^{\pi}$ according to Eq.~\eqref{linear-system-V} in Section~\ref{sec:taylored-policy-eval}. 
        \State // Policy improvement step
        \For{$i = 1, ..., N$}
            \State Update the action at the supporting state $s^i$ based on Eq.~\eqref{eq:policy-improvement}. 
        \EndFor
    \Until actions at the supporting states do not change. 
  \end{algorithmic}
\end{algorithm} 

\begin{figure*}[htbp!] \vspace{-5pt}
    \centering
    \subfloat[Kernel Taylor-based PI]{\label{fig:taylred-pi-10-10}\includegraphics[width=0.225\linewidth]{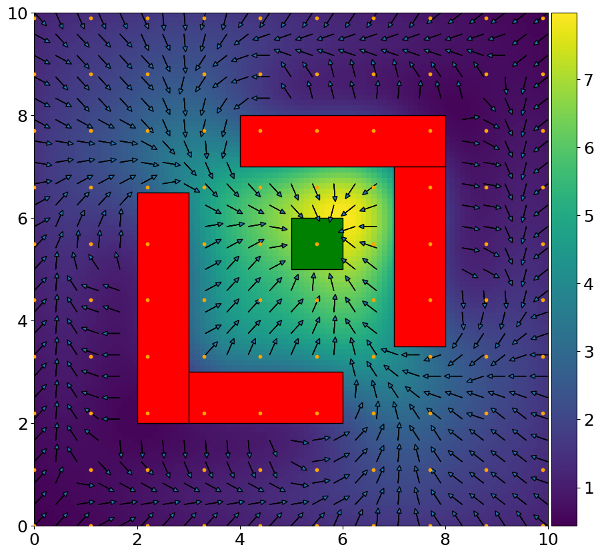}}
    \subfloat[Direct kernel-based PI]{\label{fig:closed-10x10}\includegraphics[width=0.23\linewidth]{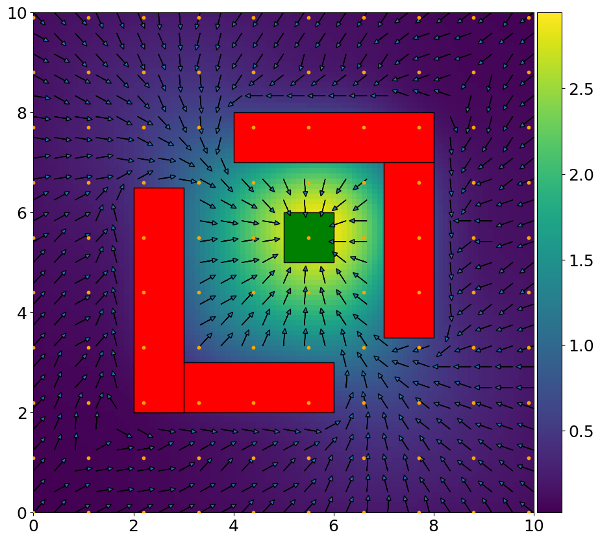}}
    \subfloat[N-FVPI]{\label{fig:nn-10x10}\includegraphics[width=0.23\linewidth]{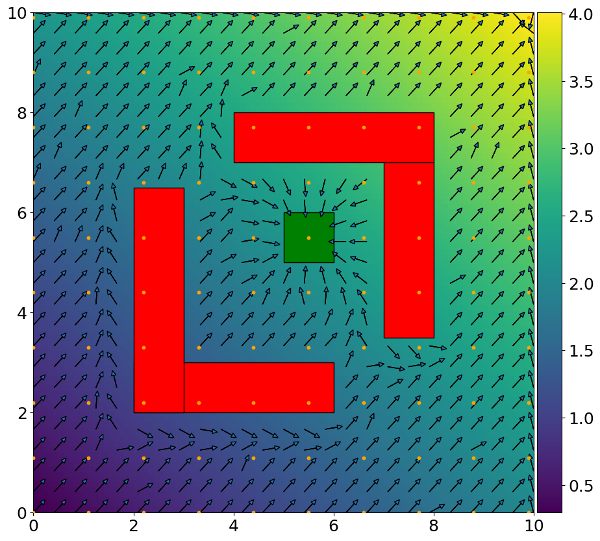}}
    \subfloat[Grid-based PI]{\label{fig:grid-10-10}\includegraphics[width=0.23\linewidth]{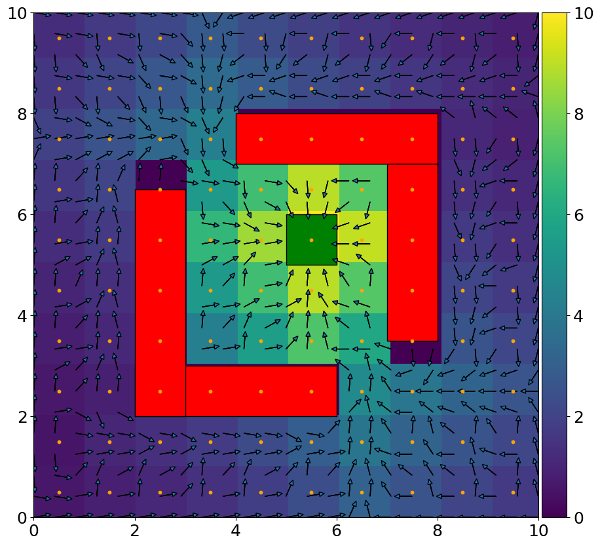}}
    \caption
    {
        Evaluation with a traditional simplified scenario where obstacles and goal are depicted as red and green blocks, respectively. 
        We compare the final value function and the final policy obtained from (a) kernel Taylor-based PI, (b) direct kernel-based PI, (c) N-FVPI, and (d) grid-based PI. 
        A brighter background color represents a higher state value.
        The policies are the arrows (vector fields), and each arrow points to some next waypoint.
        Orange dots denote the states, or the grid centers (in the case of grid-based PI), which are used to update the value functions.
    }
    \label{fig:value-fn-policy}
\end{figure*}

With the above formulations, our next step is to design an implementable algorithm that can solve the continuous-state MDP efficiently.
We extend the classic policy iteration mechanism which iterates between the policy evaluation step and the policy improvement step until convergence to find the optimal policy as well as its corresponding optimal value function.

Because our kernelized value function representation depends on the finite number of supporting states $\mathbf{s}$ instead of the whole state space, we only need to improve the policy on $\mathbf{s}$.
Therefore, the policy improvement step in the $(k+1)$-th iteration is to improve the current policy at each support state 
{
\begin{align}
     \label{eq:policy-improvement}
     \pi_{k+1}(s)&=\argmax_{a\in\mathbb{A}}\Big\{R(s, a)+\\
     &\gamma\Big((\mu_s^{a})^T\,\nabla+ 
     \frac{1}{2}\nabla\cdot \sigma_s^a\nabla\Big) v^{\pi_k}(s) \Big\}, \nonumber
\end{align}
}
where $s \in \mathbf{s}$, $\pi_{k}$ and $\pi_{k+1}$ are the current policy and the updated policy, respectively. Note that $\mu_s^a$ and $\sigma_s^a$ depend on $a$ through the transition function $p(s'|s, a)$ in Eq.~\eqref{mu-sigma-eqns}. 
Compared with the approximated Bellman optimality equation (Eq.~\eqref{diffusion-pde}), Eq.~\eqref{eq:policy-improvement} drops the term $(1-\gamma)v^{\pi_k}(s)$. This is because $v^{\pi_k}(s)$ does not explicitly depend on action $a$.
The value function of the updated policy satisfies $v^{\pi_{k+1}}(s) \geq v^{\pi_{k}}(s)$~\citep{bertsekas2012dynamic}.
If the equality holds, the iteration converges.

The final kernel Taylor-based policy iteration algorithm is pseudo-coded in Alg.~\ref{alg:kernel-based-policy-iteration}.
It first initializes the actions at the finite supporting states and then iterates between policy evaluation and policy improvement.
Since the supporting states as well as the kernel parameters do not change, the regularized kernel matrix and its inverse are computed only once at the beginning of the algorithm. 
This greatly reduces the computational burden caused by matrix inversion.
Furthermore, due to the finiteness of the supporting states, the entire algorithm views the policy $\pi$ 
as a table and only updates the actions at the supporting states using Eq.~\eqref{eq:policy-improvement}.
The algorithm stops and returns the supporting state values when the actions are stabilized. 
We can then use these state-values to get the final kernel value function that approximates the optimal solution. 
The corresponding policy for every continuous state can then be easily obtained from this kernel value function~\citep{si2004handbook}.

Intuitively, this proposed framework is efficient and powerful due to the following reasons: 
(1) by approximating the Bellman-type equation using the PDE, we eliminate the necessity in requiring a full transition function and the difficulty in computing the expectation over the next state-values; 
(2) rather than tackling the difficulties in solving the PDE, we use the kernel representation to convert the problem to a system of linear equations with characterizing values at the finite discrete supporting states. 
From this viewpoint, our proposed method nicely balances the trade-off between searching in finite states and that in infinite states. 
In other words, our approach leverages the kernel methods and Bellman optimal conditions under practical assumptions.

\section{Algorithmic Performance Evaluation} \label{Experiments}
Before testing the applicability of our proposed method in real-world environments, we conducted algorithmic evaluations in two sets of simulations in order to validate the efficiency of the proposed framework.

To do so, we eliminate the real-world complexities that are not the focus of the main method (e.g., perception, partial observability, and online re-planning). 
The first evaluation (Section~\ref{sec:first-exp}) is a goal-oriented planning problem in a simple environment with obstacle-occupied and obstacle-free spaces. 
In the second evaluation (Section~\ref{sec:second-exp}), we demonstrate that our method can be applied to a more challenging navigation scenario on Mars surface~\citep{maurette2003mars}, where the robot needs to take into account  the elevation of the terrain surface (i.e., ``obstacles" become continuous and are implicit).

\subsection{Plane Navigation}\label{sec:first-exp}

\subsubsection{Task Setup:} 
Our first test is a 2D plane navigation problem, where the obstacles and the goal area are represented in a $10m \times 10m$ environment, as shown in Fig.~\ref{fig:value-fn-policy}. 
The state space for this task is a 2-dimensional Euclidean space, i.e., $ s=[s_x, s_y]^T$ and $s \in \mathbb{S} \subseteq \mathbb{R}^2$. 
The action space is a finite set with a number of $Q$ points $\mathbb{A}(s) = \{a_i(s) | i \in \{1, ..., Q\}\}$.
Each point $a_i(s)  = [s_x + r\cos(\frac{2\pi i}{Q}), s_y+r\sin(\frac{2\pi i}{Q})]^T$ is an action generated on a circle centered at the current state with a radius $r$.
In this evaluation, we set the number of actions $Q = 12$ and the action radius as $r = 0.5 m$. 
An action point can be viewed as the ``carrot-dangling" waypoint for the robot to follow,
which is the input to the low-level motion controller.

\begin{figure}[htbp!]
    \centering
    \includegraphics[width=0.95\linewidth]{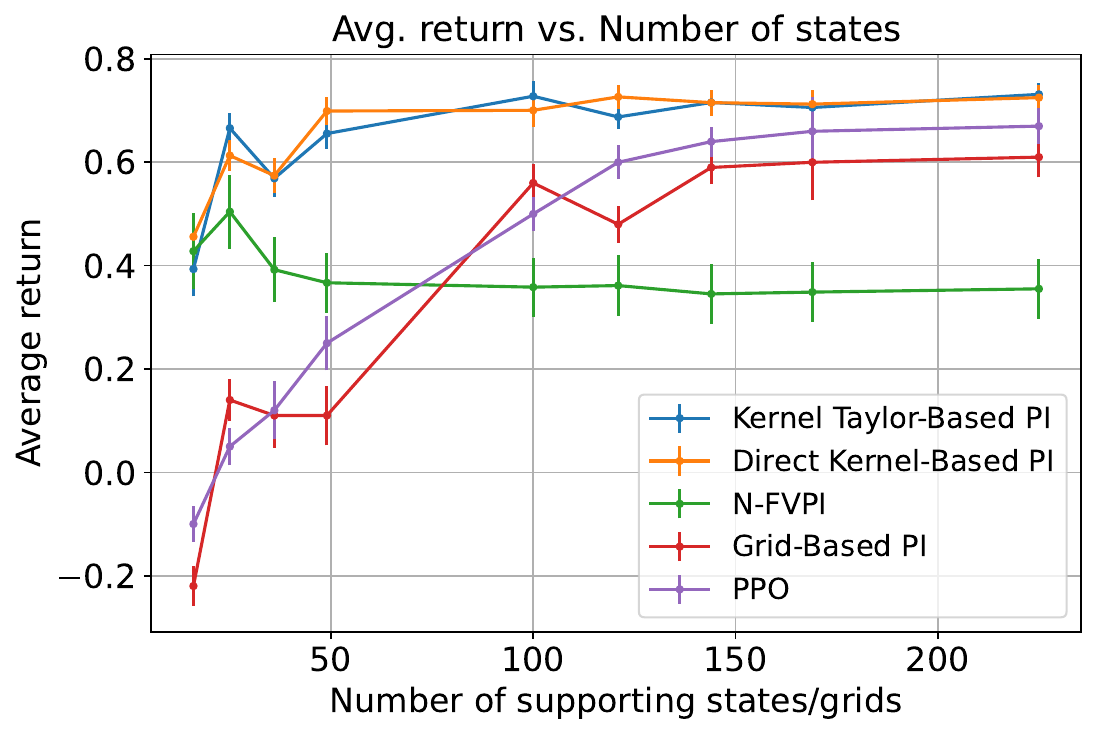}
    \caption
    { 
        The comparison of the average return of the policies computed by our method (kernel Taylor-based PI) and four other baselines.
        The x-axis is the number of supporting states/grids used in computing the policy.
        In the case of PPO, it represents the number of samples used for value and policy updates.
        The y-axis shows the average return. {The error bars represent the standard deviations.}
    }
    \label{fig:plane-nav-performance}
\end{figure}

We use a Gaussian distribution as the transition function.
The mean of Gaussian represents the selected (intended) next waypoint, while the variance is set to  $0.1 m$ 
in both $x$ and $y$ axes, accounting for the error in the low-level controller when executing the waypoint command. 
We use a sparse-reward setting, where when the agent arrives at the goal and obstacle, it receives $+1$ and $-1$, respectively.
This setting is known to be challenging for most approximate MDP methods like deep RL to solve~\citep{nair2018overcoming}.
 
Since the reward now depends on the next state, we use Monte Carlo sampling to estimate the expectation of $R(s,a)$.
The discount factor for the reward is set to $\gamma = 0.9$.
We set the obstacle areas and the goal as  absorbing states, i.e., the robot cannot transit to any other states if they are in these states.
To satisfy the boundary condition mentioned in Section~\ref{sec:bellman-eqn-pde}, we allow the robot to receive rewards at the goal state, but it cannot receive any reward if its current state is within an obstacle. 

\subsubsection{Performance measure:}
\label{sec:measure-cumulative}
Since the ultimate goal of planning is to find the optimal policy, our performance measure is based on the quality 
of the policy. 
A policy is better if it achieves a higher expected cumulative reward starting from every state.
Because it is impossible to evaluate over an infinite number of states,
we numerically evaluate the quality of a policy using the average return criterion~\citep{islam2017reproducibility}. 
In detail, 
we first uniformly sample $Q=10^4$ states to ensure a thorough performance evaluation.
Then, for each sampled state, we execute the policy for multiple trials, i.e., generating multiple trajectories, where each trajectory ends when it arrives at a terminal state (goal or obstacle) or reaches an allowable maximal number of steps. 
This procedure gives us an expected performance of the policy at any state by averaging the discounted sum of rewards over all the trajectories starting from it.
Now, we can calculate the average return criterion by averaging over the performance of sampled states.
A higher value of the average return implies that, on average, the policy gives better performance over the entire state space.

\subsubsection{Baseline Setup for Comparison:}
The performance of the proposed method is compared against four baseline approaches. 
The first baseline is \textit{direct kernel-based policy iteration}~\citep{kuss2004gaussian}. 
It approximates the value function using the kernel method with the traditional Bellman update (Eq.~\eqref{eq:value-function}) that requires the full transition probability. 
For a fair comparison, we set the kernel lengthscale as {1}, which is the same as our method. 

We also compare with the \textit{grid-based policy iteration} which is our second baseline, where the continuous state space is discretized into grids, and the vanilla policy iteration is used to solve the discretized MDP by iterating over all the grids and actions. 

The third technique, \textit{neural-fitted value function for policy iteration (N-FVPI)}, represents a class of value-based RL methods, where a neural network is used to represent the value function $v^{\pi}(s)$ to handle the continuous state space~\citep{heess2015learning}.
During the policy evaluation step, the value function's parameters are optimized to reduce the one-step squared Bellman residual via gradient descent~\citep{lutter2021value}.
Like the previous approaches, the policy is implicitly derived by selecting an action that maximizes the Bellman equation in Eq.~\eqref{bellman-equation}.
In the experiment, we use a shallow two-layer network with {100} hidden units in each layer and a SiLU activation function~\citep{elfwing2018sigmoid}.

\begin{figure*}[htbp!] 
    \centering
    \subfloat[]{\label{fig:computation-time}\includegraphics[width=0.4\linewidth]{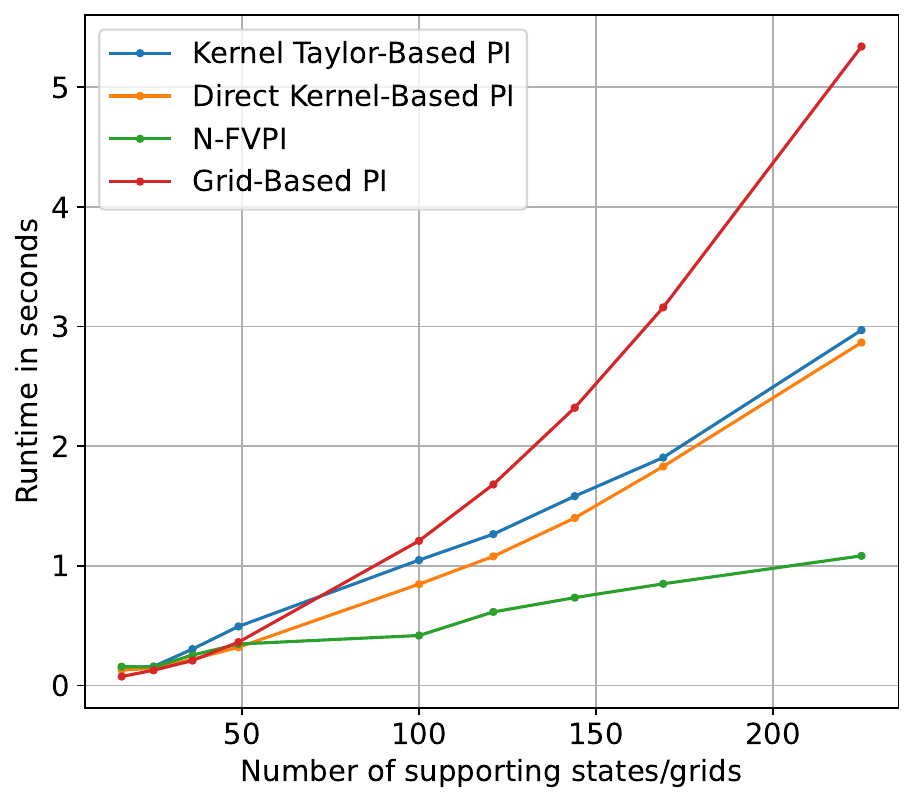}} \quad
    \subfloat[]{\label{fig:convergence-iter}\includegraphics[width=0.4\linewidth]{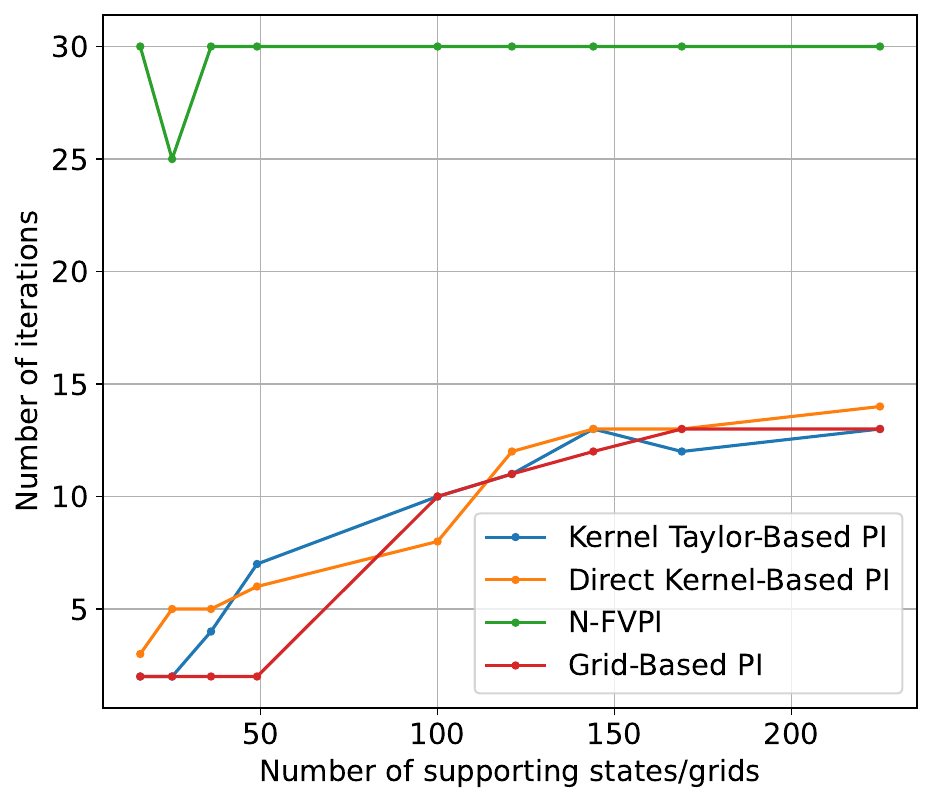}}
    \caption
    { 
        Computational time comparisons of the four value-based algorithms with changing number of states. 
        (a) The computational time per iteration.
        (b) Number of iterations to convergence.
    }
    \label{fig:computation-statistics}
\end{figure*}

To ensure a comprehensive comparison, besides the above-mentioned value-based methods, where the policy is implicitly derived from the value function, we also evaluate our method against the state-of-the-art actor-critic method, i.e., \textit{proximal policy gradient} (PPO)~\citep{schulman2017proximal}. 
It was selected as a strong baseline from the actor-critic family due to its proven track record in training complex control policies for various challenging robotics platforms, including legged locomotion~\citep{miki2022learning} and high-speed drone racing~\citep{kaufmann2023champion}. 
In the experiment, the value function and the policy are approximated using neural networks with the same configuration as N-FVPI.
Additionally, PPO generally uses zeroth-order gradients, i.e., REINFORCE-type policy gradient~\citep{sutton2018reinforcement}, to optimize the policy parameters.
In contrast, the value-based methods, i.e., grid-based, direct kernel-based, and N-FVPI, only compute the value function, and the policy is derived implicitly via the Bellman optimality equation.

\subsubsection{Results:} 

The comparison of the above methods aims at investigating three important questions: 
\begin{enumerate}
    \item In this sparse-reward navigation problem, how does the kernelized value function representation compare to alternative representations such as neural networks or grid-based methods?
    \item In contrast to our approach, the direct kernel-based method not only requires the fully known transition function but also restricts the transition to be a Gaussian distribution. Can our method with only mean and variance perform similarly to the direct kernel-based method?
    \item How does the performance of our method compare to the state-of-the-art actor-critic (reinforcement learning) method, i.e., PPO?
\end{enumerate}

The results for all five methods are shown in Fig.~\ref{fig:plane-nav-performance} on the average return with respect to the number of states used to update the value and policy functions.
The first question is answered because, among the value-based methods, the kernel method (our kernel Taylor-based and direct kernel-based PI) consistently outperforms the other two approximators (neural network and grid based).
Moreover, the second question can be answered by observing that our method has a performance as good as the direct kernel-based method which, however, requires the prerequisite full distribution information of the transition. 
This indicates that our method can be applied to broader applications that do not have complete knowledge of transition functions.
In contrast to grid-based PI, kernel-based algorithms and N-FVPI can achieve moderate performance even with
a small number of supporting states.  
It implies that the continuous representation of the value function is crucial when supporting states are sparse. 
However, increasing the number of states does not improve the performance of the N-FVPI.
Lastly, the actor-critic method, PPO, requires more samples to achieve performance on par with the value-based methods, but it exceeds the performance of the N-FVPI after using {100} samples to update the value and policy networks.
However, due to the complex approximators used (neural networks), PPO requires more samples to achieve performance on par with the kernelized value function representation.

In Fig.~\ref{fig:computation-statistics}, we compare the computational time and the number of iterations to convergence for all the value-based methods. 
The computational time of our method is less than that of the grid-based method, as revealed in Fig.~\ref{fig:computation-statistics}\subref{fig:computation-time}. 
We notice a negligible computational time difference between our and direct methods.
As a parametric method, N-FVPI has the least computational time and increases only linearly, but it does not converge as indicated by Fig.~\ref{fig:computation-statistics}\subref{fig:convergence-iter}. 

The function values and the final policies are visualized in Fig.~\ref{fig:value-fn-policy}. 
All methods except for the N-FVPI obtain reasonable approximations of the optimal value function.
Compared to our method, the values generated by the grid-based method are discrete ``color blocks".
Thus, the obtained policy is not smooth.

\subsubsection{Impact of Hyperparameters:}

\begin{figure}[bp!] 
    \centering
    \subfloat[$6\times6$]{\label{fig:taylor-pi-36}\includegraphics[width=0.485\linewidth]{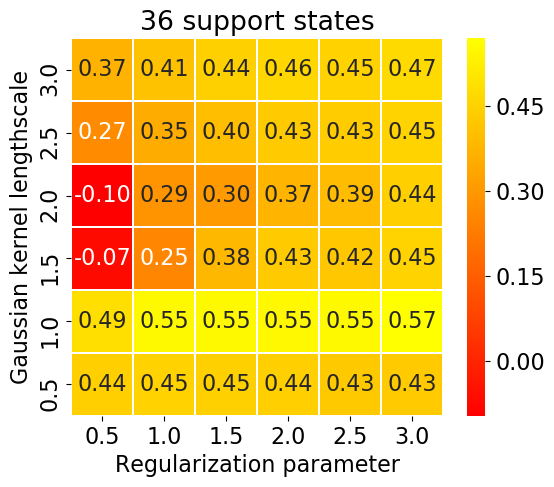}}
    \subfloat[$7\times7$]{\label{fig:taylore-pi-49}\includegraphics[width=0.5\linewidth]{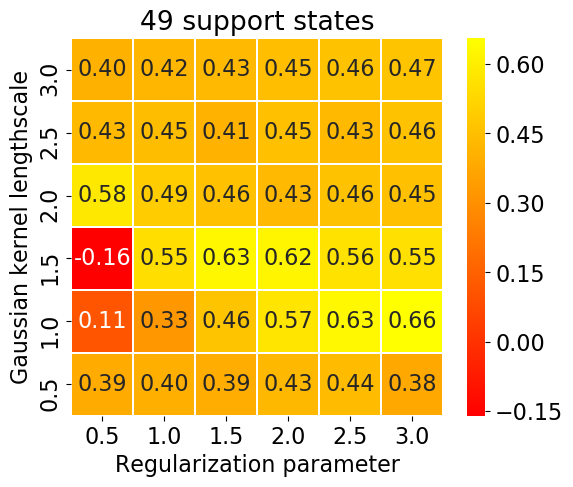}}\\
    \subfloat[$10\times10$]{\label{fig:taylor-pi-100}\includegraphics[width=0.49\linewidth]{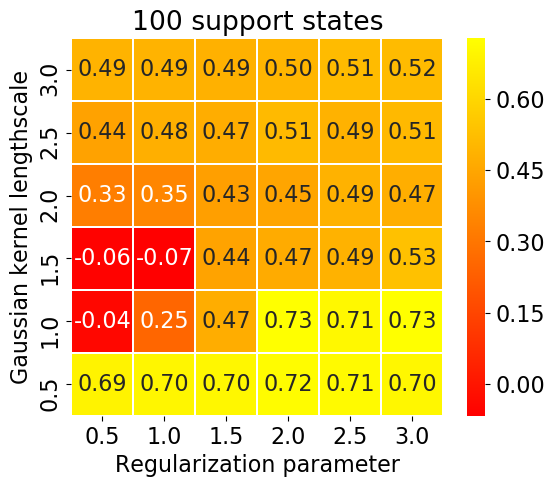}}
    \subfloat[$11\times11$]{\label{fig:taylor-pi-121}\includegraphics[width=0.49\linewidth]{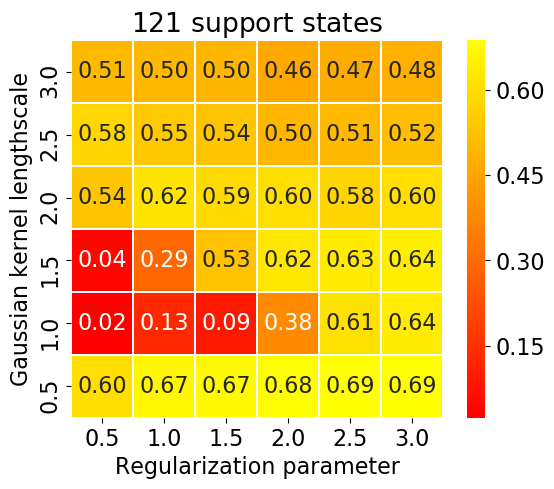}}
    \caption
    { 
        The performance matrix obtained by the hyperparameter search using (a) $6\times 6$; (b) $7 \times 7$; (c) $10\times 10$; and (d) $11 \times 11$ evenly-spaced supporting states. 
        Rows and columns represent different Gaussian kernel lengthscale and regularization parameters, respectively.
        The numbers in the color map represent the average return of the final policy obtained using the corresponding hyperparameter combination.
        The colorbar is shown on the right side of each table.
    }
    \label{fig:hyper-parameter}
    \vspace{-10pt}
\end{figure}

We also examine the impact the hyperparameters, e.g., number of supporting states and lengthscales, on the method's performance.  
To achieve this goal, we place evenly-spaced supporting states (in a lattice pattern) with different spacing resolution.
Besides the number of states, the kernel lengthscale and the regularization parameter $\lambda$ are the other two {hyperparameters} governing the performance of our algorithm. 
We present the grid-based hyperparameter search results using four different configurations of supporting states shown in Fig.~\ref{fig:hyper-parameter}.
The lengthscale and regularization parameters are searched over a set of values whose range is pre-estimated by the work-space dimensions and configurations, 
$\{0.5, 1, 1.5, 2, 2.5, 3\}$.
By entry-wise comparison among the four matrices in Fig.~\ref{fig:hyper-parameter}, we can observe that increasing the number of states leads to improving performance in general. 
However, we can find that the best performed policy is given by the $10\times 10$ supporting states configuration (Fig.~\subref{fig:taylor-pi-100}) which is not the scenario with the best spacing resolution. 
This indicates that {\em a larger number of states can also result in a deteriorating solution}, and the performance of the algorithm is a matter of {\em how the supporting states are placed (distributed)}, instead of {\em the number (resolution) of state discretization}.  
Furthermore, we can gain some insights into selecting the hyperparameters based on the number of supporting states.
Low-performing entries (highlighted with red) occur more often on the left side of the performance matrix when the number of supporting states increases.
It implies that with more supporting states, the algorithm requires a stronger regularization (i.e., greater $\lambda$ described in Section~\ref{kernel-taylored-P-E}).
On the other hand, high-performing policies (indicated by yellow) appear more at the bottom of the performance matrix when a greater number of supporting states  are present,
which means that a smaller length scale is generally required given a larger quantity of supporting states.

\subsection{Martian Terrain Navigation}\label{sec:second-exp}
\begin{figure*}[htbp!]
    \centering
    \subfloat[]{\label{fig:even-support-state}\includegraphics[width=0.23\linewidth, height=1.3in]{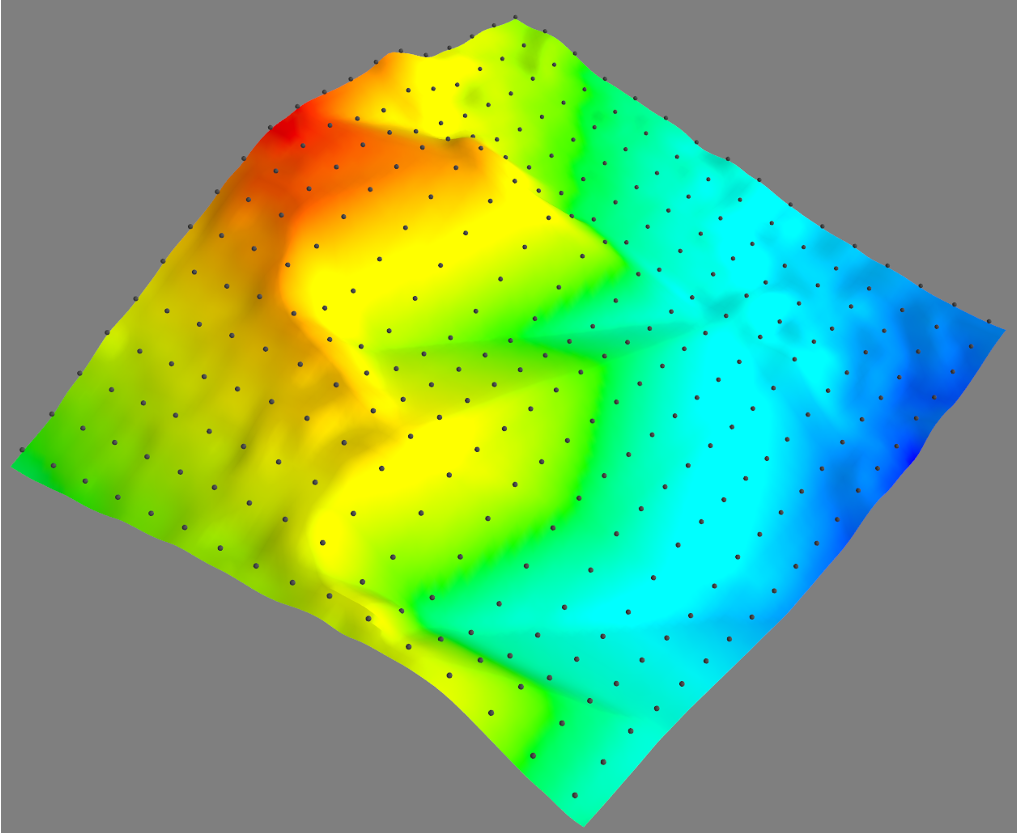}}
    \quad
    \subfloat[]{\label{fig:even-3d-policy}\includegraphics[width=0.23\linewidth,height=1.3in]{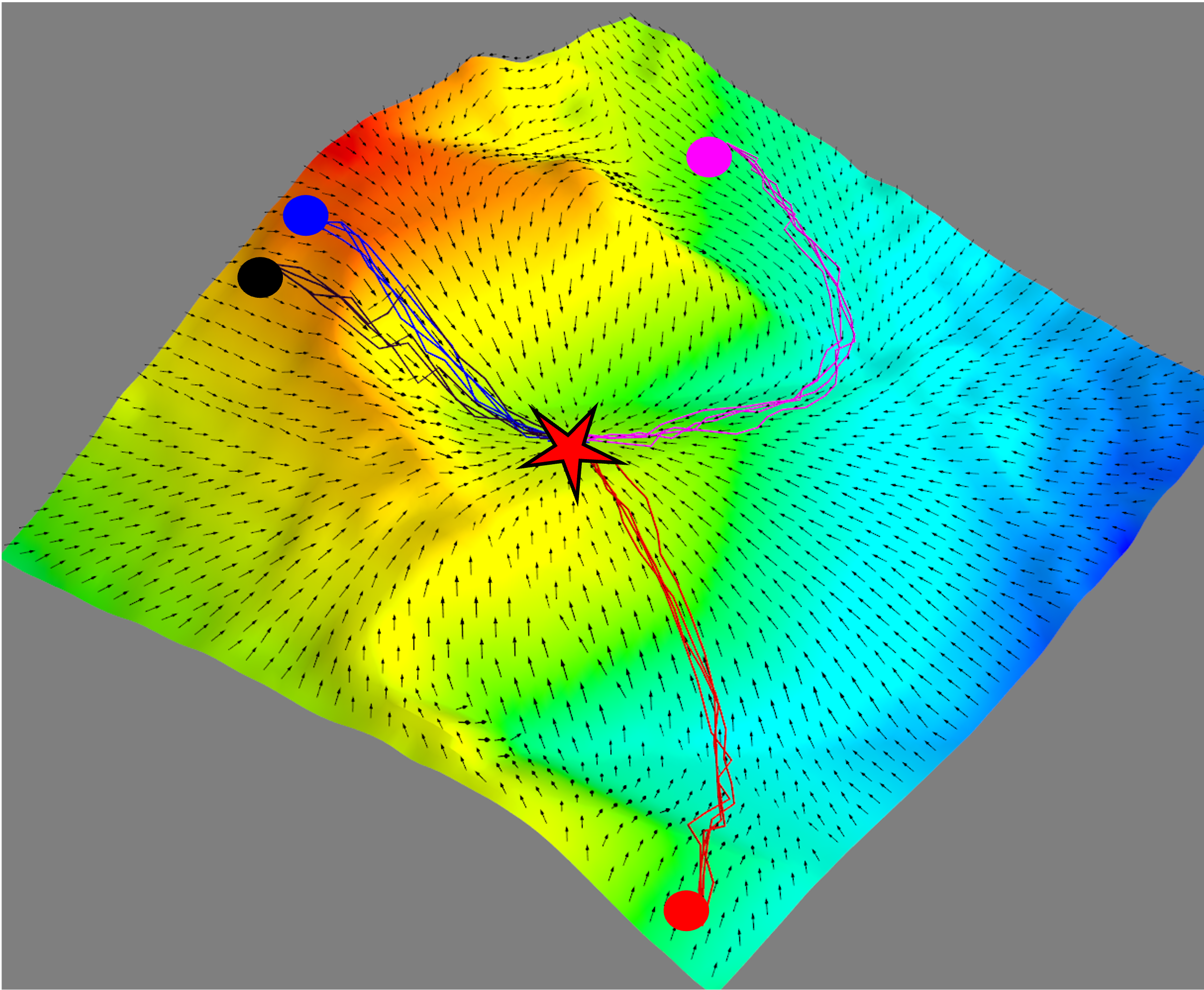}}
    \quad
    \subfloat[]{\label{fig:weighted-support-state}\includegraphics[width=0.23\linewidth,height=1.3in]{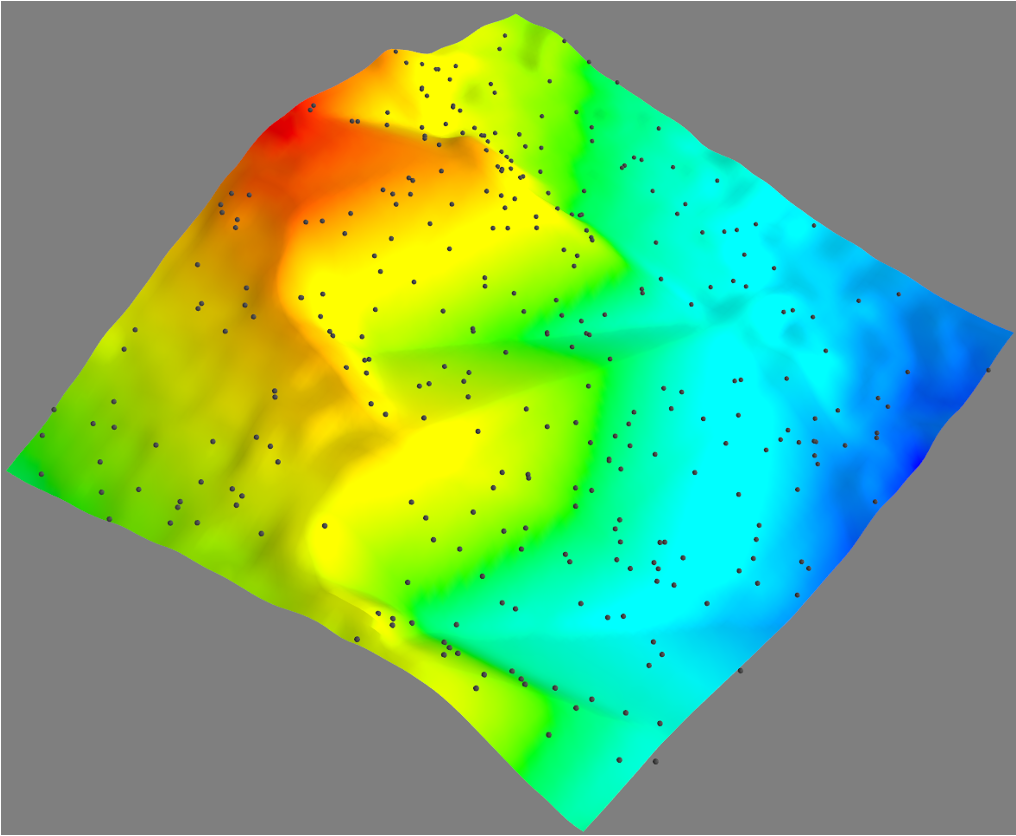}}
    \quad
    \subfloat[]{\label{fig:weighted-3d-policy}\includegraphics[width=0.23\linewidth,height=1.3in]{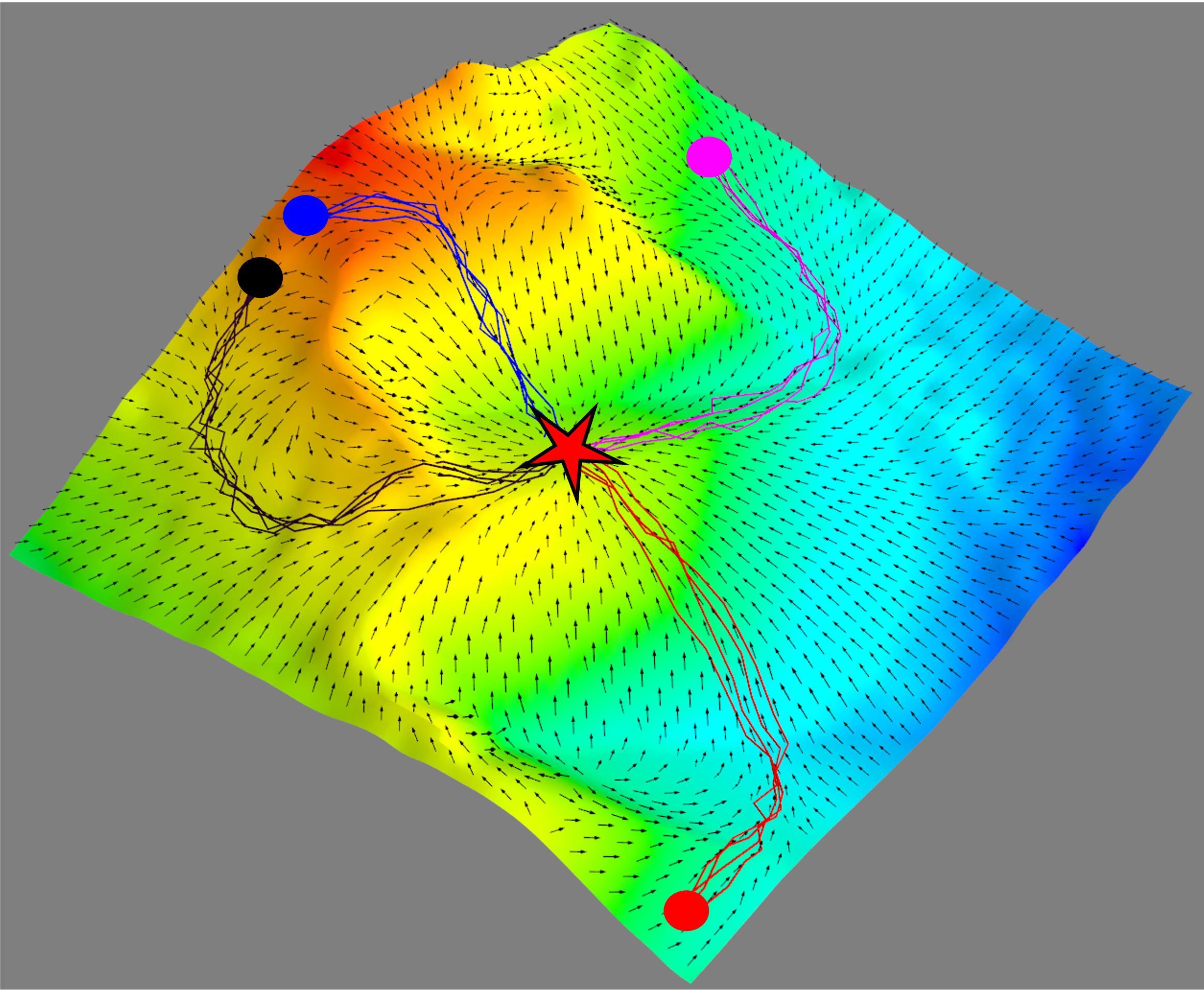}}
    \caption{ 
    Supporting state distributions and the policies for evenly-spaced selection and importance sampling-based selection.
    The 3D surface shows the Mars digital terrain model obtained from HiRISE. 
    Supporting states and policies are shown in black dots and vector fields.
    The colored lines represent sampled trajectories, which initiate from four different starting positions indicated by the circles with the same colors.
    (a)(b)~The evenly-spaced supporting states and the corresponding policy and trajectories; 
    (c)(d)~The supporting states generated by importance sampling and the corresponding policy and trajectories.   
    }
    \label{fig:policy-3d} \vspace{-10pt}
\end{figure*}

In this evaluation, we consider the autonomous navigation task on the surface of Mars with a rover.
We obtain the Mars terrain data from \textit{High-Resolution Imaging Science Experiment} (HiRISE)~\citep{mcewen2007mars}.
Since there is no explicitly presented ``obstacle", the robot only receives the reward when it reaches the goal.
If the rover attempts to move on a steep slope, it may be damaged and trapped within the same state with a probability proportional to the slope angle.
Otherwise, its next state is distributed around the desired waypoint followed by the current action.
This indicates that the underlying transition function should be the mixture of these two factors, and it is reasonable to assume that the means of the two cases are given by the current state and the next waypoint, respectively. 
We can similarly have an estimate of the variances.
The transition function's mean and variance can then be computed using the law of total expectation and total variance, respectively.

Due to the complex and unstructured terrestrial features, evenly-spaced supporting state points may fail to best characterize the underlying value function.
Also, to keep the computational time at a reasonable amount while maintaining a good performance, we leverage the importance sampling technique to sample the supporting states that concentrate around the dangerous regions where there are steep slopes.
This is obtained by first drawing a large number of states uniformly covering the whole workspace.
For each sampled state, we then assign a weight proportional to its slope angle.
Finally, we resample supporting states based on the weights. 
To guarantee the goal state to have a value, we always place one supporting state at the center of the goal area.

\begin{figure}[bp!] \vspace{-10pt}
    \centering
    \subfloat[]{\label{fig:terrain-policy-map-weighted-2d}\includegraphics[height=1.4in]{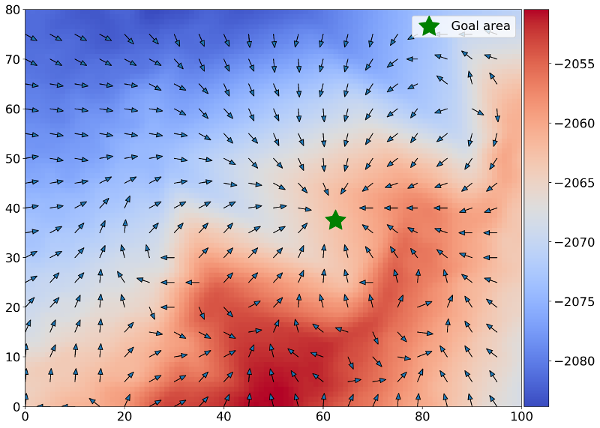}}
   \ \ 
    \subfloat[]{\label{fig:bar-chart}\includegraphics[height=1.4in]{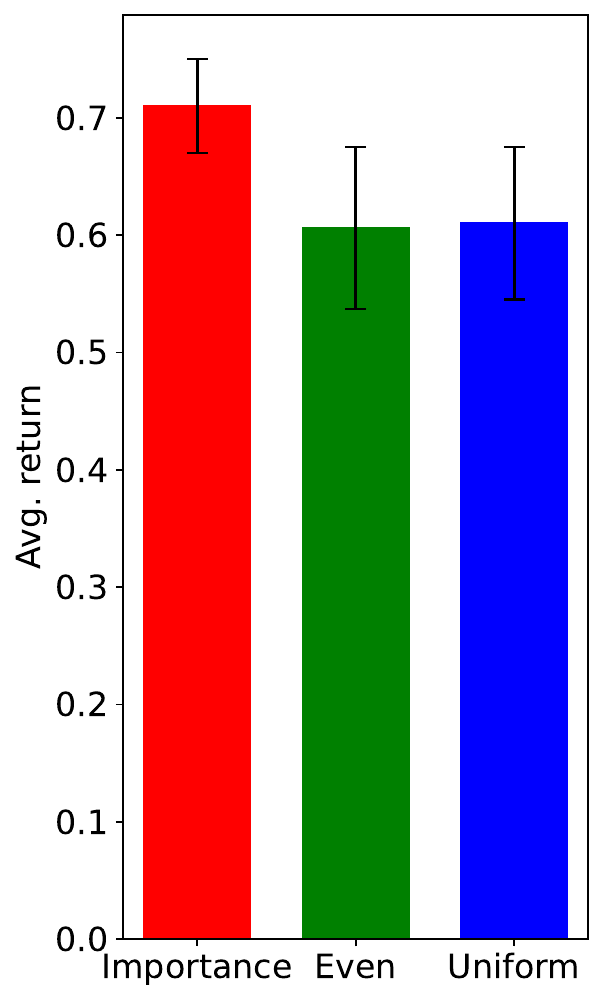}}
    \caption{
    (a) The top-down view of the  Mars terrain surface as well as the policy generated by our method with the importance sampling selection. The color map indicates the height (in meters) of the terrain.
    (b) The comparison of average returns among three supporting state selection methods using the same number of states. Red, green, and blue bars indicate the performance of importance sampling selection, evenly-spaced selection, and uniform distribution sampling selection, respectively. 
    }
    \label{fig:policy-2d} \vspace{-10pt}
\end{figure}

Fig.~\ref{fig:policy-3d}\subref{fig:even-support-state} and Fig.~\ref{fig:policy-3d}\subref{fig:weighted-support-state} compare the two methods of differing supporting state selections.
The supporting states given by the importance sampling-based method are dense around the slopes.
These supporting states better characterize the potentially high-cost and dangerous areas than the evenly-spaced selection scheme.  
We selected four starting locations from which the rover needs to plan paths to reach a goal location. 
For each starting location, we conducted multiple trials following the produced optimal policies.
The 
trajectories generated with the importance sampling states 
in Fig.~\ref{fig:policy-3d}\subref{fig:weighted-3d-policy} attempt to approach the goal (red star) with minimum distances, and at the same time, avoid dangerous terrains by choosing more leveled/even surfaces. 
In contrast, the trajectories obtained using the evenly-spacing states in Fig.~\ref{fig:policy-3d}\subref{fig:even-3d-policy} approach the goal in a  more aggressive manner which can be risky in terms of safety. 
It indicates that a good selection of supporting states can better capture the state value function and thus produce finer solutions. 
This superior performance can also be reflected in Fig.~\ref{fig:policy-2d}\subref{fig:bar-chart}.
The policy obtained by the uniformly sampled states shows similar performance to the one generated by the evenly-spacing states, both of which yield smaller average return than the importance-sampled case. 
A top-down view of the policy is shown in Fig.~\ref{fig:policy-2d}\subref{fig:terrain-policy-map-weighted-2d} where the background color map denotes the elevation of terrain.

\section{Autonomy System Design}\label{sec:system}

Previous algorithmic evaluations assume that a well-defined MDP can be abstracted and constructed from the real-world problem and focus on algorithmic property/performance assessments.
However, in reality, specifying an MDP for stochastic motion planning problems requires a known task region for constructing the state space as well as detailed information on the environment for transition and reward functions. 
These requirements are generally very difficult to satisfy if we deploy the robots into the real world, especially in the off-road environment where the task region is hard to define and no prior map is provided. 
In such cases, the robot needs to use its onboard sensors to acquire information and make decisions online.
Specifically, it needs to process the observations, construct a new MDP within the observed area, compute the corresponding policy, and execute it. 
This process should repeat until the task is completed. 
In this paper, we only focus on generating a local policy within the observable area.
Global planning in any unknown environment might require information-driven (active) sensing to achieve a trade-off between the exploration of unknown space and the exploitation of observed space, which, however, is not the focus of this work.   

Furthermore, we confine the MDP construction process to only relying on the environments' geometric information and develop the navigation system based on it.
This capability is vital if only perception depth information is available (e.g., with point cloud from any ranging sensors) through which we can leverage mostly the reconstructed geometry.
As illustrated in Fig.~\ref{fig:mars}, the uneven terrains pose many challenges (e.g., traversals of hills, ridges, valleys, slopes, etc.) as one of the primary features affecting the robot's motion efficiency is geometry. 

\begin{figure*}[htbp!]
    \centering
     \includegraphics[width=0.8\linewidth]{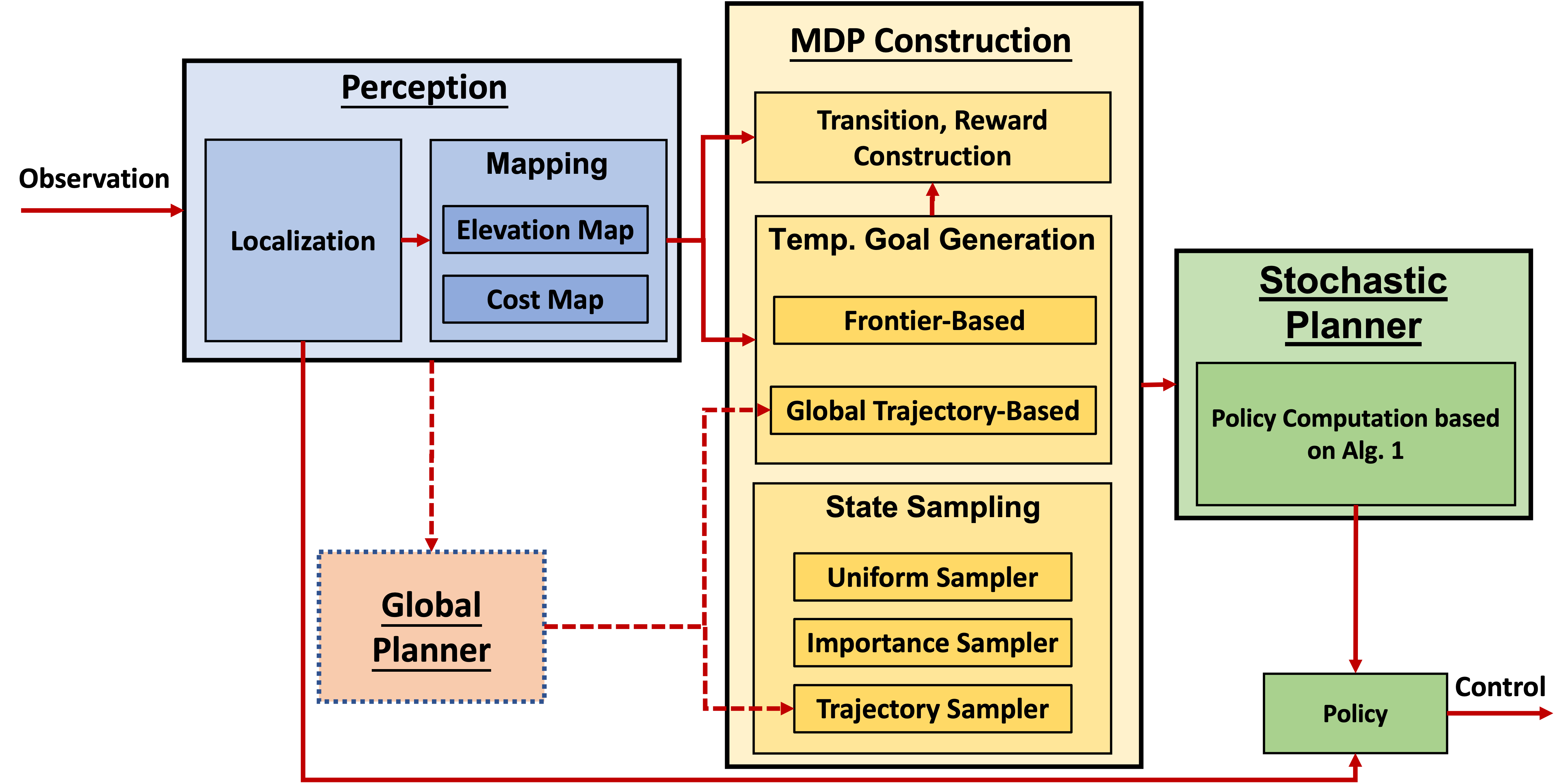}
     \centering
    \caption{An overview of the system for navigation in prior-unknown environments.
    The red arrows represent the direction of the information flow.
    The global planner module in the dashed block is optional. 
    }
    \label{fig:system}
\end{figure*}

In this section, we present our autonomy system that extracts the MDP elements (supporting states, transition function, and reward function) from depth sensor data and connects the perception to the action loop. 
Fig.~\ref{fig:system} provides an overview of the system. 

\subsection{Perception}
The perception module is responsible for processing sensor observation and providing information for planning. 
We use LiDAR to acquire geometric information about the environment.
Based on the point cloud, it generates the robot state (pose) and a map of the environment in the field of view. 
The pose information can be estimated by any existing LiDAR localization methods (e.g., ~\citet{bresson2017simultaneous}).
For the map representation, we utilize the cost map to provide occupancy information for identifying obstacle-free regions for navigation.
However, there exist scenarios in uneven terrains where it might inaccurately model the {\em navigability} of certain regions. 
For instance, areas with considerable elevation might be misrepresented as occupied, despite 
the possible traversability due to a non-steep gradient of the surface.
Thus, we use a $2.5D$ elevation map~\citep{fankhauser2018probabilistic} to provide a finer depiction of the terrain shape.
Combined with the variance design based on the slope information described in the previous section, the robot can reason about navigability beyond mere occupancy.

\subsection{MDP Construction}\label{subsec:mdp-construction}
The second part of the system builds an MDP using the mapping information.
In the following, we explain this construction process for each MDP element.

\subsubsection{State and Action Spaces:}
The state and action spaces are pre-defined with the given  vehicle type (ground).
Specifically, the state is represented as the robot pose with position and orientation information $(x, y, \theta)$,  and the action consists of linear and angular speeds $(v, \omega)$. 
Since our method deals with the discrete action space, we can tessellate the continuous action into evenly-spaced intervals (i.e., equi-distanced linear/angular speed levels). 
The discretization resolution, as well as the minimum and maximum values for the speeds, depend on the robot's capability and the task's complexity.

\subsubsection{Reward Function:}
When using the cost map, the reward function consists of two parts $R(s) = (1-\mathbb{I}_{s_g}(s)) c(s) + r_g \mathbb{I}_{s_g}(s)$.
The first part $c(s)$ is the obstacle penalty provided by the cost map, $\mathbb{I}_{s_g}(s)$ is an indicator function for checking whether the state belongs to the goal region $s_g$, and $r_g$ is the goal reward. 
If only the elevation map is used, the obstacle penalty is not included, and the robot only receives a positive reward if it arrives at the goal.
Since we focus on planning within the observable space only, we need to generate a dummy goal region $s_g$ in the space of the field of view.  
In other words, it needs to set a temporary goal if the final goal region is outside of the robot's current observed area.
We use two heuristic approaches to generate this temporary goal.
The first one is inspired by the frontier-based concept which generates a temporary goal chosen at the boundary of the current observable map and with the least distance towards the final goal~\citep{burgard2005coordinated, yamauchi1997frontier}.
The second method utilizes the path generated by a global planner if available (usually just for high level guidance).
The temporary goal can be determined by ``cropping" a path segment from the global path within the field of view only, where the path segment starts from the current robot pose and extends to a pre-defined length along the cropped path, 
and the final pose on the extracted path segment serves as the temporary goal.
Note that such a global planner is not mandatory for our method as it can be replaced with any frontier-based goal selection method.

\subsubsection{First Two Moments of the Transition Function:}
We assume that the robot motion is generated based on $s' = f(s, a, h) + \epsilon$, where $f(s, a, h)$ is a deterministic function representing the discrete-time motion model related to vehicle dynamics, e.g., a differential drive model; $h(s)$ is the slope angle at state $s$ derived from the elevation map; and $\epsilon$ is a noise term independent of the state and action. 
If the elevation map is used, the motion model considers the effect of the elevation on the robot's motion. 
Otherwise, we treat $h(s)=0$ (flat surface) for all states in the current state space. 
To construct $f(s, a, h)$, we modify the Dubin's car model to take into consideration of the terrain slope, $x' = (\frac{\pi}{2} - h(s))(x + v\Delta t \cos{\theta}) + h(s)x$,  $y' = (\frac{\pi}{2} - h(s))(y + v\Delta t\sin{\theta}) + h(s)y$, and $\theta' = \theta + \omega \Delta t$.
Here, $h(s) \in [0, \frac{\pi}{2}]$ is the slope angle at state $s$ and $\Delta t$ is the time discretization resolution.
Intuitively, this model penalizes the distance traveled on the surface with a larger slope angle. 
Based on this formulation and Eq.~\eqref{mu-sigma-eqns}, the first and second moments can be computed as $\mu(s,a,h) = \Delta_a s + \mathbb{E}[\epsilon]$ and $\sigma(s,a,h) = \mathbb{V}[\epsilon] + \mu(s,a,h)\mu(s,a,h)^T$, where $\Delta_a s = f(s,a,h) - s$ is the state shift after applying action $a$.
In this work, we choose the mean $\mathbb{E}\left[\epsilon\right]=0$ and variance proportional to the slope $\mathbb{V}\left[\epsilon\right]=kh(s)$, where $k \geq 0$ is a constant that modulates the impact of the slope on the model's uncertainty.
Specifically, a larger value of $k$ indicates increased uncertainty in the robot's motion on elevated terrains.
This variance formulation allows the planner to be more cautious about navigating on high-slope terrain surfaces, enabling the robot to avoid these hazardous areas.
The advantage of our method is obvious from the above modeling perspective.
Since we only need to model the mean and variance of the noise $\epsilon$, it is not necessary to acquire the exact probability distribution of the noise.

\subsubsection{Supporting States:}\label{sec:system.support_states}
The last part of the construction is state sampling, which is responsible for distributing the supporting states within the state space.
The positions and the number of supporting states are critical as they determine the accuracy of the value function and also computational time.
More supporting states generally provide a better estimation of the optimal value function but require more time to compute as shown in Section~\ref{sec:first-exp}. 
We consider uniform sampling, importance sampling, and trajectory sampling strategies in this work.
All these sampling methods require a region to distribute the supporting states.
The first two methods introduced in Section~\ref{sec:first-exp} and Section~\ref{sec:second-exp} need  prior knowledge of the environment and specification of a sampling region, e.g., a rectangular workspace.
In greater detail, the uniform sampling tessellates a pre-defined region into equal-sized cells and uses the cell vertices as the supporting states. 
The importance sampler first uniformly samples a large number of states and then selects them based on some weighting criteria, e.g., the slope of the terrain at a given state point. 
The trajectory sampler utilizes the global path generated by a global planner as a heuristic to define the planning and sampling region. 
It extracts a path segment on the global path starting from the current robot pose and distributes the state samples around this path segment, whose length can also be determined by the maximum linear speed of the robot. 
It distributes the state samples around the same path segment extracted from the global trajectory-based goal generation method.
This method is especially useful when fast online computation is required since it does not need to search the entire planning region. 

\subsection{Policy Computation and Execution}
With all the MDP elements ready, we can use Alg.~\ref{alg:kernel-based-policy-iteration} to compute a policy for the constructed MDP.
Then, this policy is used for generating control actions $a = (v, \omega) = \pi(s)$.
It is necessary to discuss two motion planning and control strategies of this system based on the global planner's ability to plan an executable path on uneven terrains.

\subsubsection{Periodic Replanning:}
\begin{figure*}[htbp!]
    \centering
    \subfloat[The top-down view of the environment]{\label{fig:topdown-sim}\includegraphics[width=2.1in, height=1.5in]{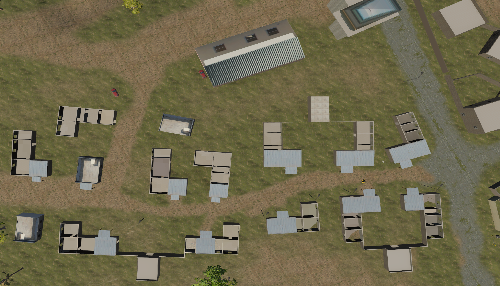}}
    \hspace{0.2cm}
    \subfloat[A close view of navigable space in the simulator]{\label{fig:sim}\includegraphics[width=2.1in, height=1.5in]{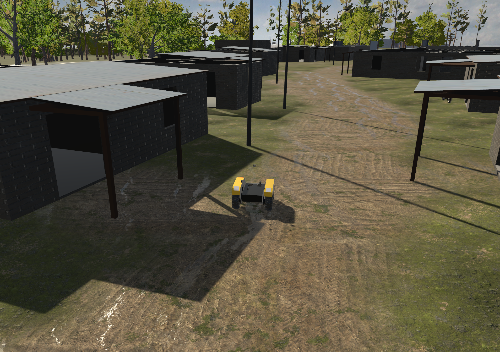}}
    \hspace{0.2cm}
    \subfloat[An example of the narrow alleyway in the environment]{\label{fig:topdown-sim}\includegraphics[width=2.1in, height=1.5in]{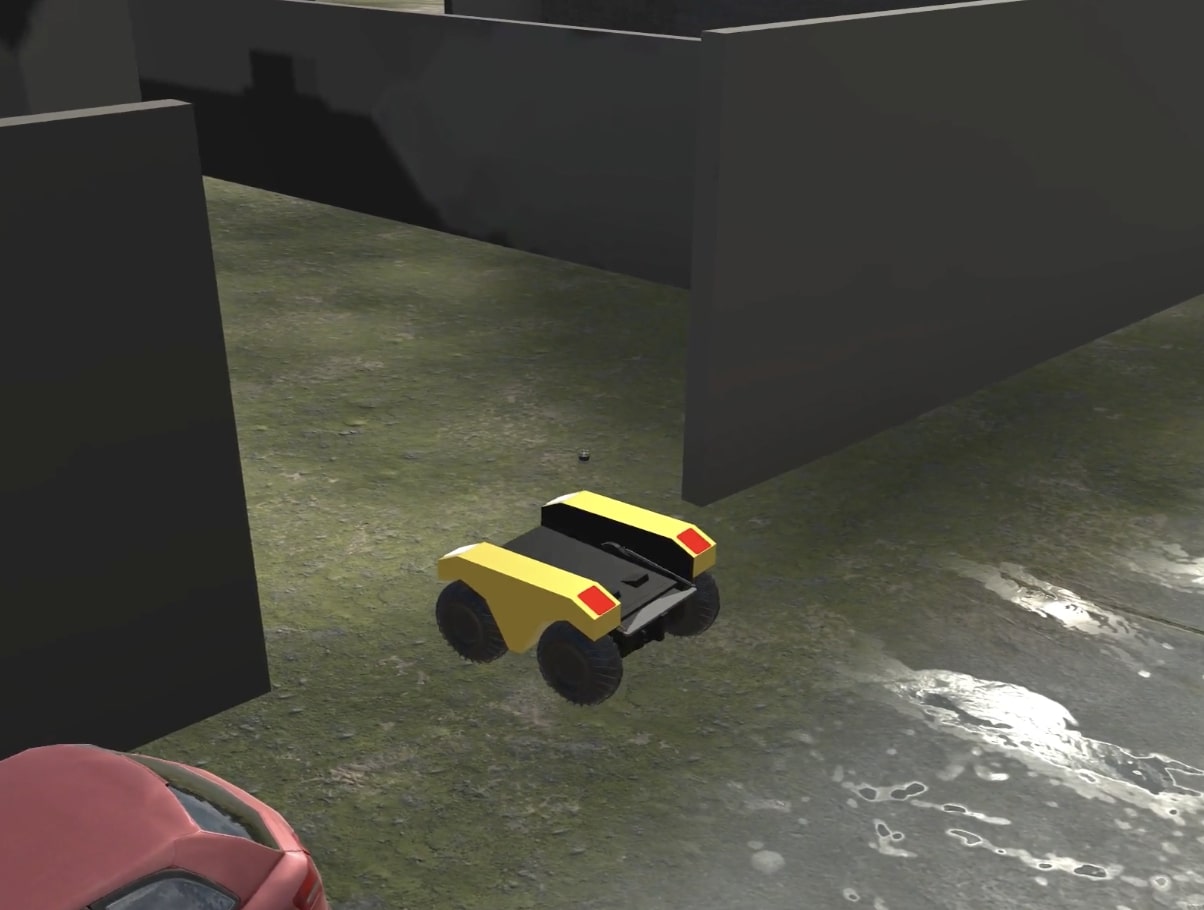}}
    \caption{The Unity simulation environment.}
    \label{fig:high-fidelity}
\end{figure*}

Conventional commonly-used global path planners (e.g., sampling-based or graph-based methods) generally do not consider the tracking ability of the lower-level planner and controller.
As a result, when the low-level controller cannot execute the planned path accurately, the robot may deviate and fail to complete the task.
This issue is particularly prevalent when the robot navigates on highly unstructured uneven terrains, where behaviors such as slippage or other unpredictable motion outcomes can occur frequently.
In contrast to the traditional path planning framework that calculates a path while disregarding the inherent uncertainty when executing the global path, the proposed approach computes a feedback policy across a specified region by considering the possible errors that the low-level controller can easily make due to possibly fast-varying terrain elevations. 
In essence, our proposed method synthesizes the entire process of planning and control under uncertainty within a single framework.

We use the frontier-based exploration method~\citep{yamauchi1997frontier} to generate temporary goal points. 
It is worth mentioning that when the planning region is large, a high computational load can naturally occur, resulting in a pausing behavior when the robot needs to replan. This can be mitigated and tuned by reducing the cropped planning region. 
Once the policy is computed within the selected region, it can query the action anywhere in the defined state space in real-time, as action computation only requires iterating over a finite number of actions.
Also, replanning is typically invoked periodically, initiated either when the robot reaches a predetermined interim goal or when new map information necessitates modifying the current policy, such as when the initially planned policy leads to a collision. 
Upon either event, the robot ceases its current policy and recalculates a new one incorporating this updated information. 
While this approach might cause delays due to re-planning over a large region, it facilitates effective robotic operation in situations where the path determined by the conventional global planner is difficult for the low-level controller to track.

\subsubsection{Real-Time Planning with a Global Trajectory:}
If the global planner can provide a reasonable trajectory, which incurs only moderate tracking error, this information can reduce the search space by using the trajectory sampling strategy to obtain states around the globally-planned path.
The region encompassed by these states may be conceptually viewed as a tube within which the robot must remain, similar to the funnels constructed in~\citep{majumdar2017funnel}.
Focusing the policy search around the global path allows MDP construction and policy iteration to be performed in real-time. 
Subsequently, this affords the implementation of the Model Predictive Control (MPC) paradigm for real-time execution. 
After each policy iteration computation, the robot executes a singular action from the feedback policy and initiates a re-planning process.

\section{Experiments Using a Realistic Physics Simulator}\label{sec:lidar-exp}

\subsection{Experimental Setup}

To accomplish the 
autonomous navigation system described above, we first integrate the perception module and test the system's performance.
The robot is no longer provided with a prior map of the environment,  and it can observe the environment and obtain state information from its onboard depth sensor.

\begin{figure*}[htbp!] 
    \centering
    \subfloat[{10} Waypoints]{\label{fig:market-loop}\includegraphics[width=2.1in, height=1.5in]{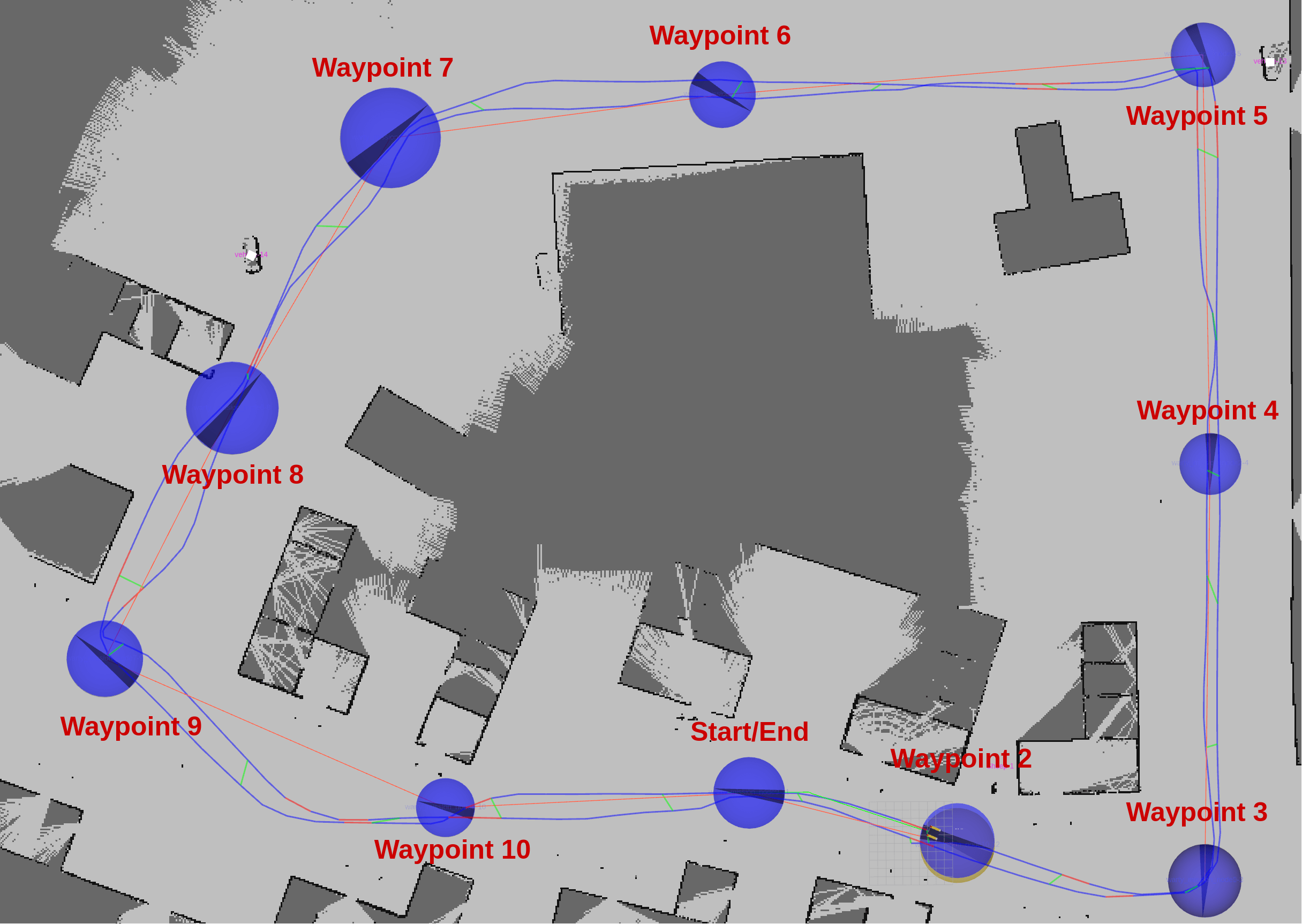}}
    \hspace{0.2cm}
    \subfloat[{5} Waypoints]{\label{fig:market-loop}\includegraphics[width=2.1in, height=1.5in]{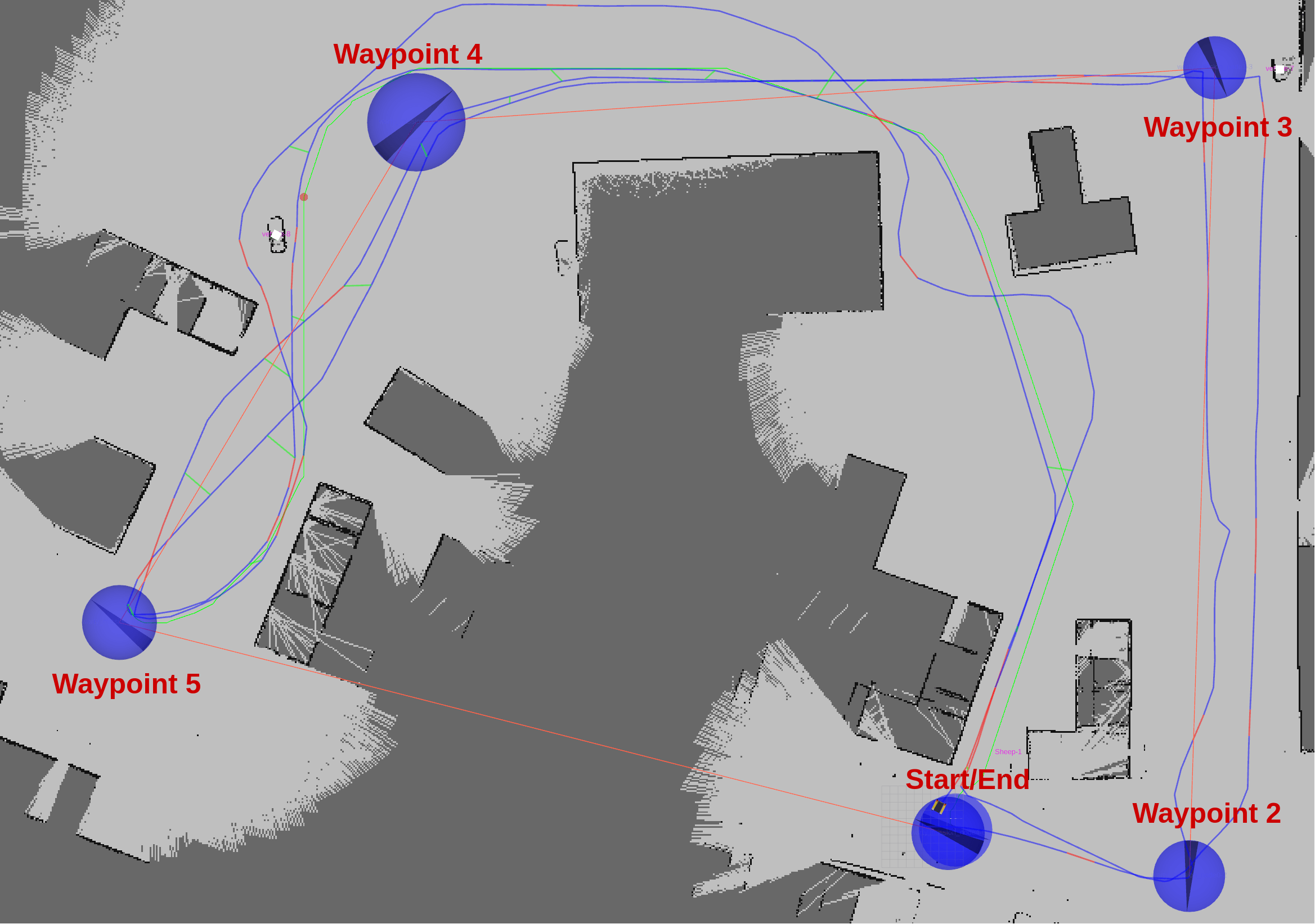}}
    \hspace{0.2cm}
    \subfloat[{4} Waypoints]{\label{fig:market-loop-hard}\includegraphics[width=2.1in, height=1.5in]{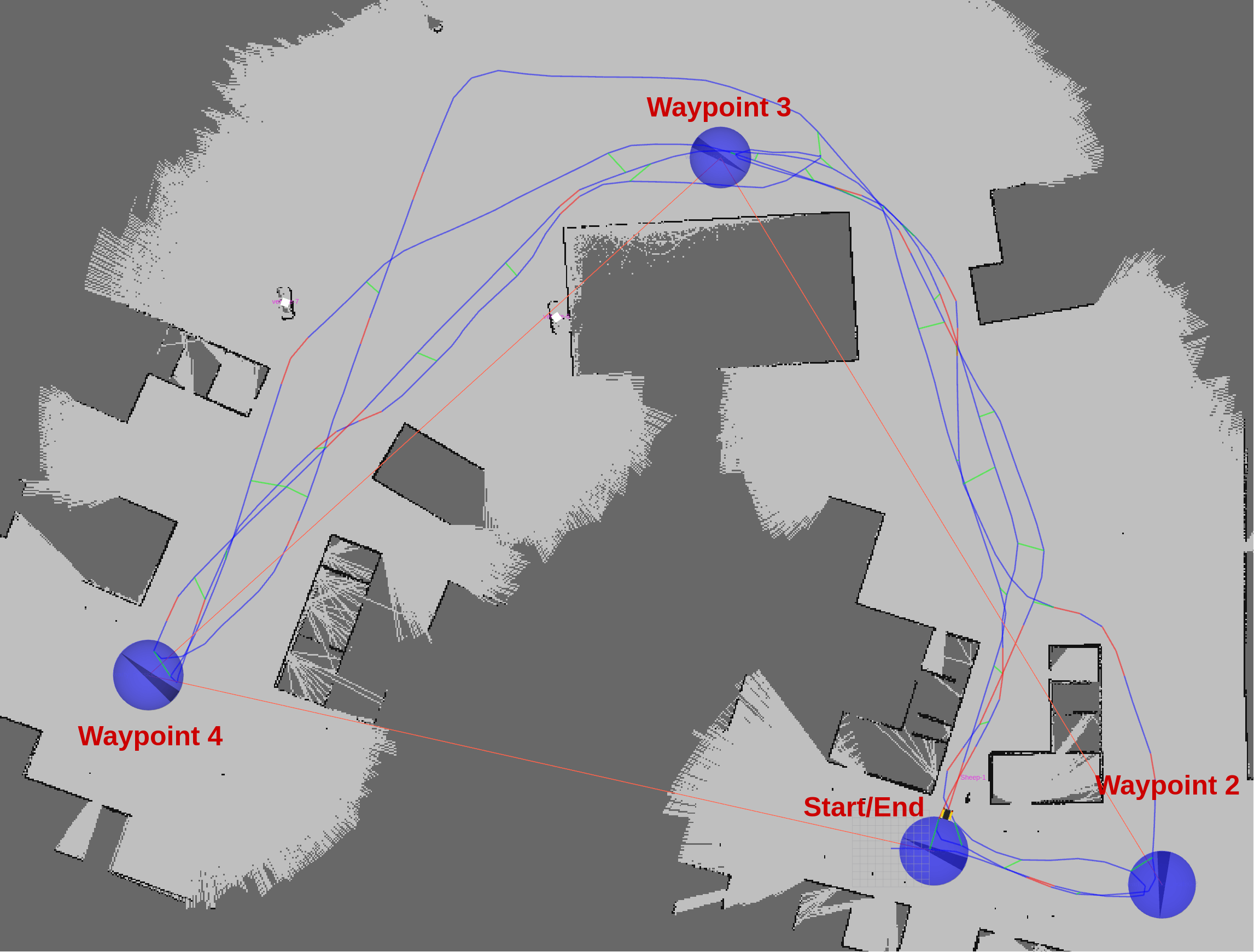}}
    \caption
    { 
    The mission maps and sampled robot trajectories for each waypoint setting.
    The blue circles are the waypoints.
    For each setting, the robot starts from the initial position, navigates sequentially to the next points, and returns to the initial position. 
    The blue paths represent the trajectories in two runs using our method. 
    The straight red lines between waypoints show the mission's connection (i.e., order of waypoints).
    The other colored short path segments are related to robot's simultaneous localization and mapping (SLAM).
    In particular, the short red line segments scattered along the traveled path correspond to the pose graph factors after the successful execution of Iterative Closest Point (ICP) registrations~\citep{grisetti2010tutorial}. 
    Instances of successful loop closures are demonstrated by green lines linking two navigated paths. 
    Lastly, the extended light green lines in (a) and (b) represent the global plan. } 
    \label{fig:mission-maps}
\end{figure*}

Our test scenario is designed to assess the navigation efficiency of the robot within a high-fidelity Unity simulation environment, where the robot is required to navigate without a prior map. 
The simulator simulates a ClearPath Warthog differential drive ground vehicle with a {64}-beam LiDAR.
Its task is to navigate to a set of pre-defined waypoints sequentially and return to the initial position.
The snapshot of the simulator environment and a bird-eye map view can be observed in Fig.~\ref{fig:high-fidelity}.
This simulation environment encapsulates a range of challenging characteristics typically present in real-world off-road settings. 
Among these challenges are the uneven and textured ground terrain, clusters of densely packed obstacles, and narrow alleyways, which all pose substantial complexities for effective navigation. 
In addition, our mission definitions require the robot to plan and navigate areas without clearly distinguishable dirt-roads.
These aspects of the environment necessitate a planning and motion control method that is precise enough to navigate constricted passages and robust enough to accommodate disturbances arising from uneven surfaces.

We set the robot's minimum and maximum values for linear and angular speeds as $v_{min} = -1m/s$, $v_{max} = 1m/s$, $\omega_{min} = -1.5 rad/s$ (turning right), and $\omega_{max} = 1.5 rad/s$ (turning left).
In this experiment, we test the real-time local planning capability of the system, where an anytime version of A* (ATA*) \citep{likhachev2005anytime} is employed as the global planner to provide the desired reference trajectory, and we sample support states around it, as described in Section~\ref{sec:system.support_states}.

We compare our method with  the nonlinear model predictive control (NMPC) technique for path optimization. NMPC is a deterministic local trajectory optimization approach (see \citet{allgower2012nonlinear} for an overview), and it 
 has a track record of documented successes in path planning over real-world challenging terrains~\citep{gregory2016application, howard2007optimal}. The NMPC is still 
benchmarking for highly agile trajectories in cluttered and challenging environments (e.g., drone flight at speeds up to 20 m/s~\citep{sun2022comparative}). 
The traditional three-layer planning and control system often adopts this pipeline. 
We implemented the NMPC which produces smooth trajectories that respect the robot's dynamics by optimizing a cost function penalizing the path deviation from a desired global trajectory using algorithms from the NLOPT library~\citep{nlopt}.

The two approaches were tested under three settings: {10}, {5}, and {4} waypoints, respectively. 
Fig.~\ref{fig:mission-maps} shows the waypoint positions, the constructed occupancy map, and two sampled trajectories for each setting using our method. 
The action space for our method is discretized evenly into $8 \times 8$ intervals which are dense enough to approximate continuous actions.
For both methods, we choose $6m$ as the length of trajectory sampling, meaning that the robot plans $2s$ in the future if it uses the maximum speed $3m/s$.

\begin{table}[h]
  \centering
  \caption{Comparison of Diffusion MDP local planner (our method) and NMPC in different scenarios.
  Best-performing metrics are highlighted in \textbf{bold}.
  The statistics are averaged over $10$ runs.}
  \resizebox{\columnwidth}{!}{%
    \begin{tabular}{cccccc}
    \toprule
     & {10} Waypoints & {5} Waypoints & {4} Waypoints \\
    \midrule
    \multicolumn{1}{c}{\underline{\textbf{Diffusion MDP}}} \\
    Tracking Error ($m$) & $0.43 \pm 0.09$ & \bm{$0.56 \pm 0.14$} & \bm{$0.51 \pm 0.11$} \\
    Avg. Jerk ($m/s^3$) & $6.73 \pm 0.86$ & \bm{$6.35 \pm 0.62$} & \bm{$6.57 \pm 0.34$} \\
    Completion Time ($s$) & \bm{$157.2 \pm 1.3$} & \bm{$198.6 \pm 11.2$} & \bm{$177.8 \pm 9.3$} \\
    Success Rate & \bm{$10/10$} & \bm{$10/10$} & \bm{$10/10$} \\
    \midrule
    \multicolumn{1}{c}{\underline{\textbf{NMPC}}} \\
    Tracking Error ($m$) & \bm{$0.41 \pm 0.14$} & $0.63 \pm 0.11$ & $0.69 \pm 0.16$ \\
    Avg. Jerk ($m/s^3$) & \bm{$5.9 \pm 0.56$} & $6.77 \pm 0.41$ & $7.14 \pm 0.39$ \\
    Completion Time ($s$) & $161.5 \pm 4.7 $ & $204.6 \pm 4.1$ & $203.3 \pm 3.5$ \\
    Success Rate & \bm{$10/10$} & {8}/{10} & {4}/{10} \\
    \bottomrule
    \end{tabular}%
    }
  \label{tab:comparison}
\end{table}%

\subsection{Metrics}

Both methods are evaluated with the following four metrics:
\begin{itemize}
    \item \textit{Tracking Error}: This metric measures the averaged distance between the global path and the executed path in the $(x,y)$ axes. 
    A smaller tracking error indicates the method can better execute the plan with a smaller deviation from the global path and is preferred. 
    \item \textit{Average Jerk (Avg. Jerk)}: This metric indicates the robot's average sum of acceleration changes along  $(x,y)$ directions across a mission.
    Generally, a smaller value reflects a smoother motion execution of the mission and is preferred.
    \item \textit{Completion Time}:  
    The completion time is the total traversal time to arrive at all the designated waypoints. 
    It is computed over the successful runs only.
    \item \textit{Success Rate}: 
    The success rate indicates the number of times the robot arrives at all the waypoints out of {10} runs without collision or getting stuck. 
\end{itemize}
Compared to the averaged cumulative rewards that we used for evaluating the basic algorithmic performance (Section~\ref{sec:measure-cumulative}), these metrics are more relevant to the robot navigation task.
We record these metrics only during the execution of the optimized trajectory (or policy), which excludes the global planner's influence on the evaluation.

\subsection{Result}
\begin{figure*}[htbp!] 
    \centering
    \subfloat{\includegraphics[width=0.31\linewidth,height=1.35in]{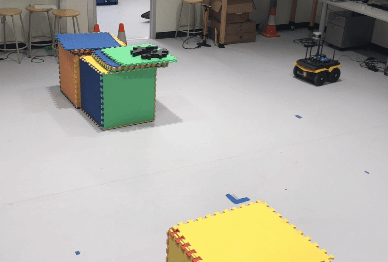}}
    \hspace{0.15cm}
    \subfloat{\includegraphics[width=0.31\linewidth,height=1.35in]{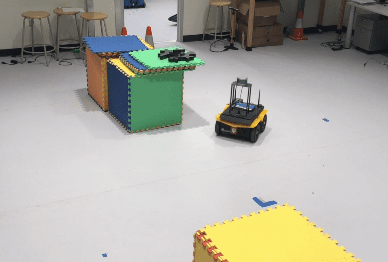}}
    \hspace{0.15cm}
    \subfloat{\includegraphics[width=0.31\linewidth,height=1.35in]{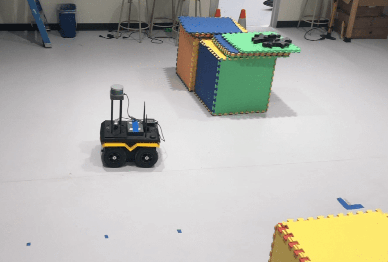}}\\
    \subfloat{\includegraphics[width=0.31\linewidth,height=1.35in]{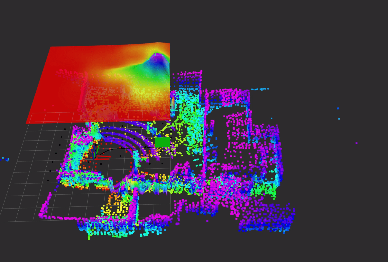}}
    \hspace{0.15cm}
    \subfloat{\includegraphics[width=0.31\linewidth,height=1.35in]{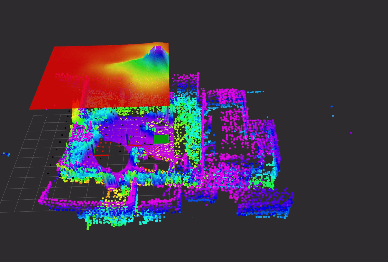}}
    \hspace{0.15cm}
    \subfloat{\includegraphics[width=0.31\linewidth,height=1.35in]{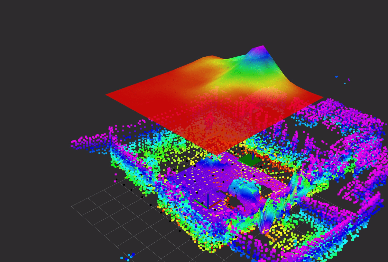}}\\
    \subfloat{\includegraphics[width=0.31\linewidth,height=1.35in]{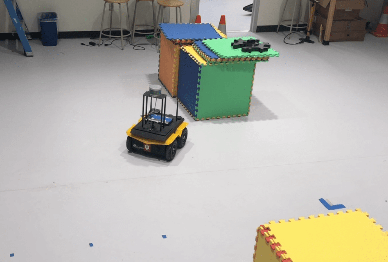}}
    \hspace{0.15cm}
    \subfloat{\includegraphics[width=0.31\linewidth,height=1.35in]{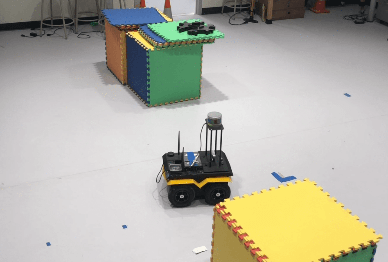}}
    \hspace{0.15cm}
    \subfloat{\includegraphics[width=0.31\linewidth,height=1.35in]{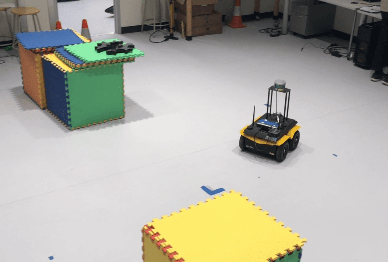}}\\
    \subfloat{\includegraphics[width=0.31\linewidth,height=1.35in]{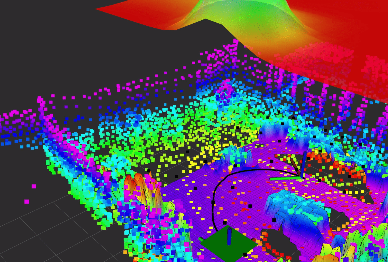}}
    \hspace{0.15cm}
    \subfloat{\includegraphics[width=0.31\linewidth,height=1.35in]{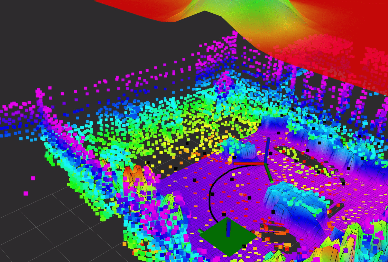}}
    \hspace{0.15cm}
    \subfloat{\includegraphics[width=0.31\linewidth,height=1.35in]{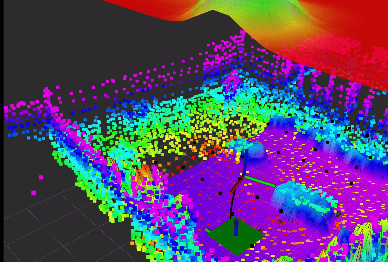}}\\
    \caption
    { 
    The indoor environment where our method is tested for validating its basic obstacle-avoidance and goal arrival behaviors.
    The first and third rows show the vehicles' real-world behaviors at different timesteps, and the second and fourth rows depict the corresponding visualization results.
    The task for the robot is to navigate through the gap between  boxes (approximately $1.5m$) and arrive at the green square shown in the second row, and then return back to the initial position. 
    The colored dots are the point cloud generated from the LiDAR sensing.
    Upper and lower $2.5D$ color maps represent the value function and the elevation map, respectively.
    The height of the value function represents the state value.
    The planned expected trajectories are shown in black curves. 
    The small axes represent the pose of the robot.}
    \label{fig:indoor-exp}
\end{figure*}

The result is shown in Table~\ref{tab:comparison}.
In general, our method is more efficient and can complete the task using a shorter traversal time than NMPC. 
Additionally, our method completes all {10} runs for the three waypoint settings, while NMPC  achieves $100\%$ success rate only in the {10} waypoint setting (i.e., with sufficient waypoint guidance). 
Since the NMPC method attempts to follow the global path exactly, the trajectory optimized by NMPC sometimes oscillates around and overshoots the global path.
This oscillation causes the vehicle to reduce its speed due to frequent turning.
In contrast, since our method considers the uncertainty (via the second moment) of the robot's motion, it can reason about a region around the global path. 
As a result, as long as the robot remains within the area supported by the sampled states, it does not show oscillation behavior.  
The second row (average jerk) of the table can reflect this comparison, where in the {5} and {4} waypoint scenarios, our method produces a smaller jerk, resulting in smoother trajectory execution, while NMPC needs to stop abruptly when it overshoots far away from the global path. 
We can also observe that the global path overshooting problem of NPMC leads to larger tracking errors in the last two waypoint settings.

It is also worth pointing out that the robot uses a shorter time to complete the task in the {4} waypoint mission than in the {10} waypoint. 
This can be explained by observing the trajectories the robot generated in the two missions in Fig.~\ref{fig:mission-maps}.
Although the {10} waypoint mission provides denser waypoints, the robot is restricted to following these predefined points, which are not necessarily the shortest route.
In contrast, the {4} waypoint mission has less restriction on which trajectory the robot should follow. 
As a result, the robot chooses a shortcut to navigate from Waypoint {2} to Waypoint {3} as shown in Fig.~\ref{fig:mission-maps}\subref{fig:market-loop-hard}.

\section{Real-World Demonstrations in Indoor Cluttered and Outdoor Unstructured Environments}\label{sec:real-world-exp}
To demonstrate the applicability of our system in the real world, we conducted extensive real-world trials in different scenarios.
The goal is to validate whether the proposed method can be used to generate effective policies that navigate a ground vehicle in complex and unstructured environments.
The videos of the robot behaviors are demonstrated in Extensions 1 and 2. 

\subsection{Hardware Setup}
We implement our system on two ground vehicles, ClearPath Jackal and Husky.
Since Jackal is small in dimension and has a small wheel traction, it is only used for indoor experiment.
It is equipped with a 16-beam Velodyne LiDAR for localization and mapping, a quad-core 2.7GHz CPU and 16GB RAM for the onboard computation.
Husky is used for navigating in outdoor environments because it is larger in size and provides more traction when moving on rough terrains.
Husky is equipped with a 64-beam Ouster LiDAR for more detailed terrain mapping and a Lord Microstrain 3DM 
IMU for more accurate localization outdoor.
It has an eight-core 2.7GHz Intel I7 CPU and 64GB RAM for the onboard computation.

\subsection{Indoor Cluttered Environment}
The setup is an indoor $5m^2$ environment with random boxes as obstacles, and the task is to navigate the ClearPath Jackal to a goal location and return to the start position.
The first aim is to validate the basic obstacle avoidance behavior in a-prior unknown environment.
The second aim is demonstrating that our system can achieve efficient navigation behavior without a global planner.
We leverage the elevation representation for the MDP construction.
Traditionally, the obstacle penalty in the reward (or cost) function is used as an indirect method for expressing the potential consequences of collisions.
In this experiment, we also demonstrate that we can achieve the same obstacle-avoidance behavior without the obstacle penalty, using only the two statistical moments of the robot's motion based on the elevation map (see Section~\ref{subsec:mdp-construction}). 
This prepares us for navigation on uneven terrains demonstrated in the next section, where obstacle boundaries may be unidentifiable and the traditional use of the cost map may be inadequate.

Since the planning region is pre-specified\footnote{Note that it does not mean that the environment map needs to be provided a-prior.}, we evenly distribute $6$ supporting states in each of the $x,y$ dimensions, so that the distance between two neighboring states is $1m$ in each dimension. 
However, because it is difficult to predict what heading angles the robot may take during the task execution, we need supporting states to cover one full rotation of the heading angle, i.e.,$[-\pi, \pi)$. 
Specifically, we place $10$ equidistant states in the $\theta$ dimension so that the distance between two neighboring states along the $\theta$ dimension is $\frac{\pi}{5}$.
The total number of states is $360$.

The range of linear and angular speeds are set to $v \in [-0.5m/s, 1.5m/s]$ and $\omega \in [-1.5rad/s, 1.5rad/s]$, respectively.
The environment, the robot's motion sequence, and detailed planning information are shown in Fig.~\ref{fig:indoor-exp}.
The initial position of the robot is placed in front of the obstacles, and the goal is between the two obstacles and is indicated by the green square on the second row of Fig.~\ref{fig:indoor-exp}.
Initially, due to the occlusion by randomly placed boxes, the robot can only build a partial elevation map using the observed points. 
Then, it uses this initial map to calculate the value function within the state space.
The value function is shown as the upper $2.5D$ surface in the second row.
To aid the visualization, we only draw the value function over the position variables at the current robot's angle, and the height of the surface represents the state value. 
We can observe that the value function correctly characterizes the environment's geometry and the task.
Specifically, the goal region has the highest state value.
Since the policy always selects actions that maximize the state value, executing the policy from this value function will allow the robot to converge at the goal point.
Additionally, on the value function surface, the narrow ridge between the two obstacles reflects the navigable space correctly.
As a result, the policy derived from the value function can navigate the passage safely, as shown in the first row of Fig.~\ref{fig:indoor-exp}. 
As the robot executes its planned policy, it receives more sensing points and gradually builds a more detailed elevation map.
This progressive refinement of the elevation map can be seen from the second and fourth rows.
After the robot arrives at the goal, it re-plans and returns to the initial position, which is shown in the third row. 
As the robot's forward maximum speed ($v_{max} = 1m/s$) is larger than its backward maximum speed ($v_{min} = -0.5m/s$), it strategically reverses and aligns its forward axis with the initial position to ensure leveraging the maximum linear speed (the behavior can be observed in the supplementary videos).
This process shows that the computed value function allows the robot to reason about its under-actuated dynamics through the two moments.
This also reveals that our method exhibits a {\em speed-adaptive} behavior for the most efficient motion.

\subsection{Outdoor Unstructured Environment}\label{sec:outdoor-exp}
\begin{figure*}[htbp!]
    \centering
    \subfloat{\includegraphics[width=0.24\linewidth,height=1.1in]{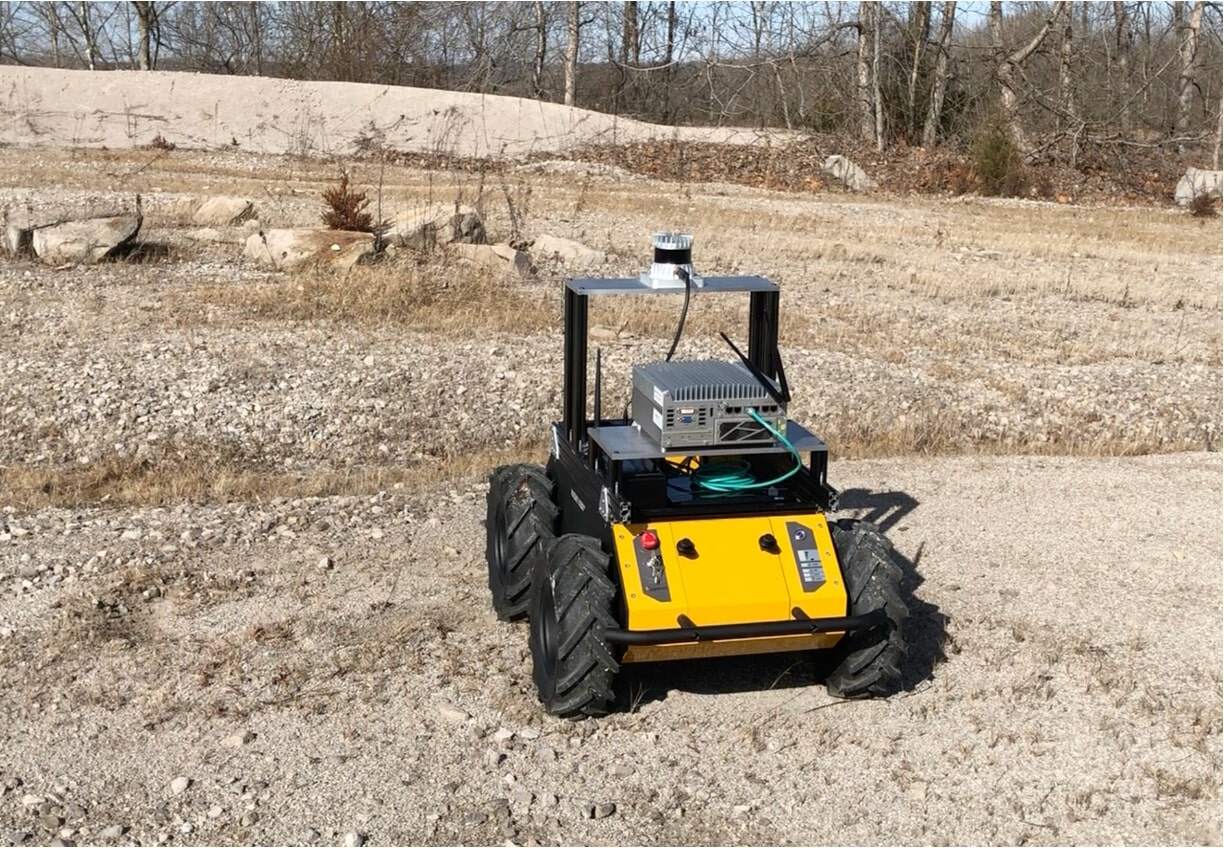}}
    \hspace{0.1cm}
    \subfloat{\includegraphics[width=0.24\linewidth,height=1.1in]{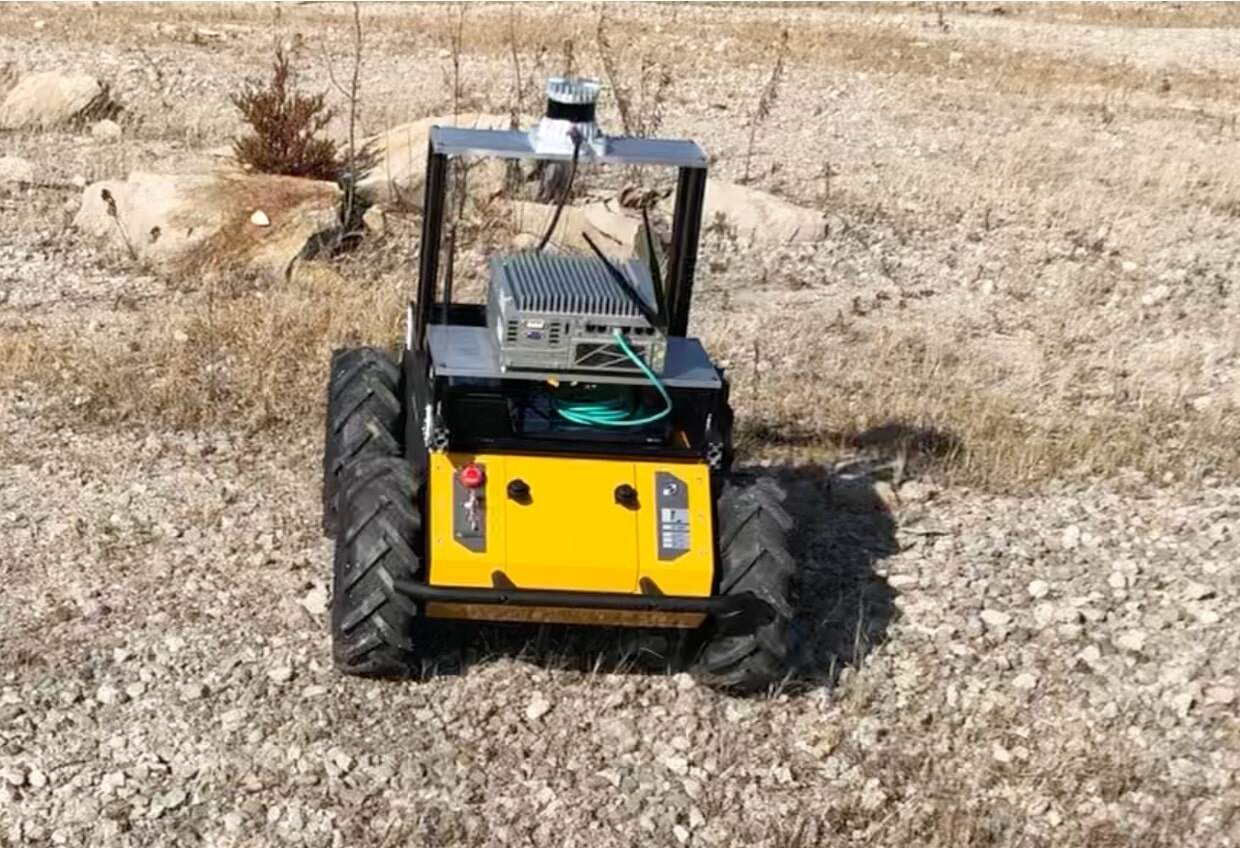}}
    \hspace{0.1cm}
    \subfloat{\includegraphics[width=0.24\linewidth,height=1.1in]{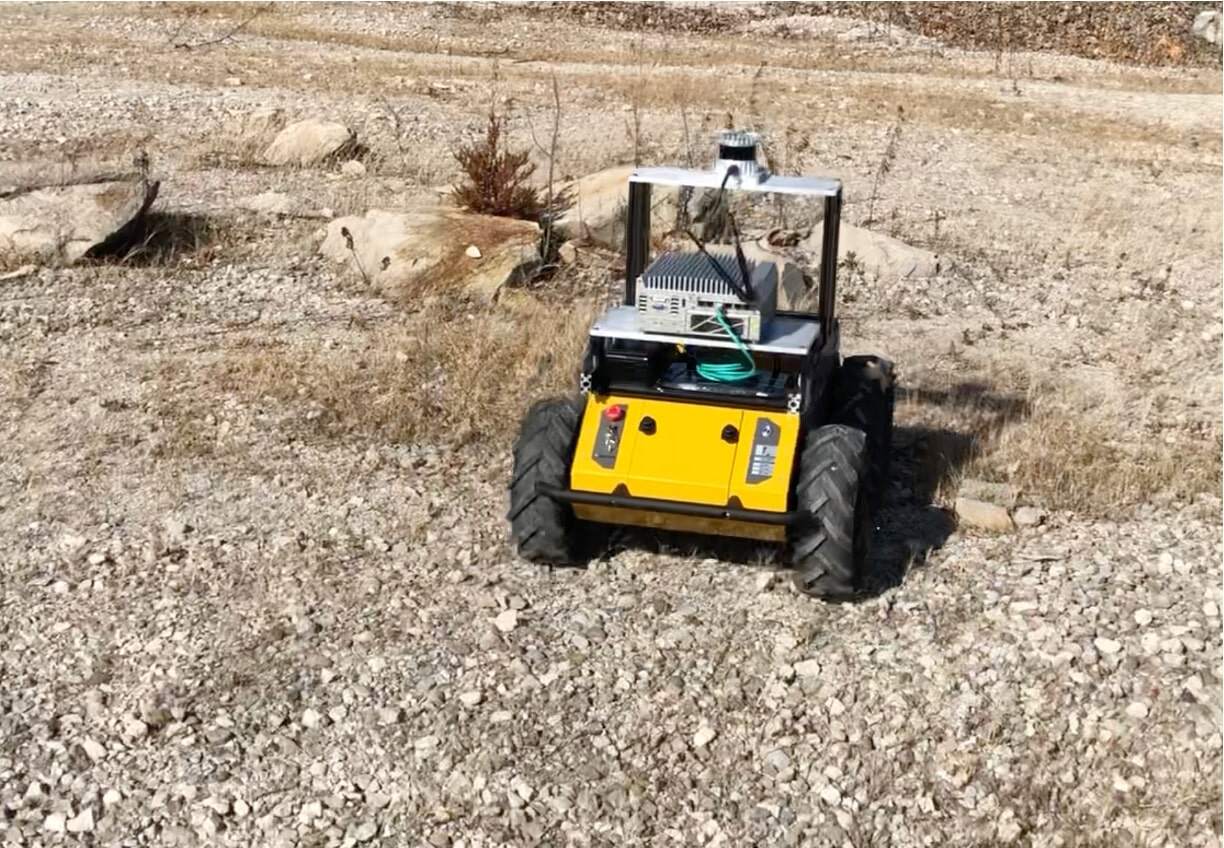}}
    \hspace{0.1cm}
    \subfloat{\includegraphics[width=0.24\linewidth,height=1.1in]{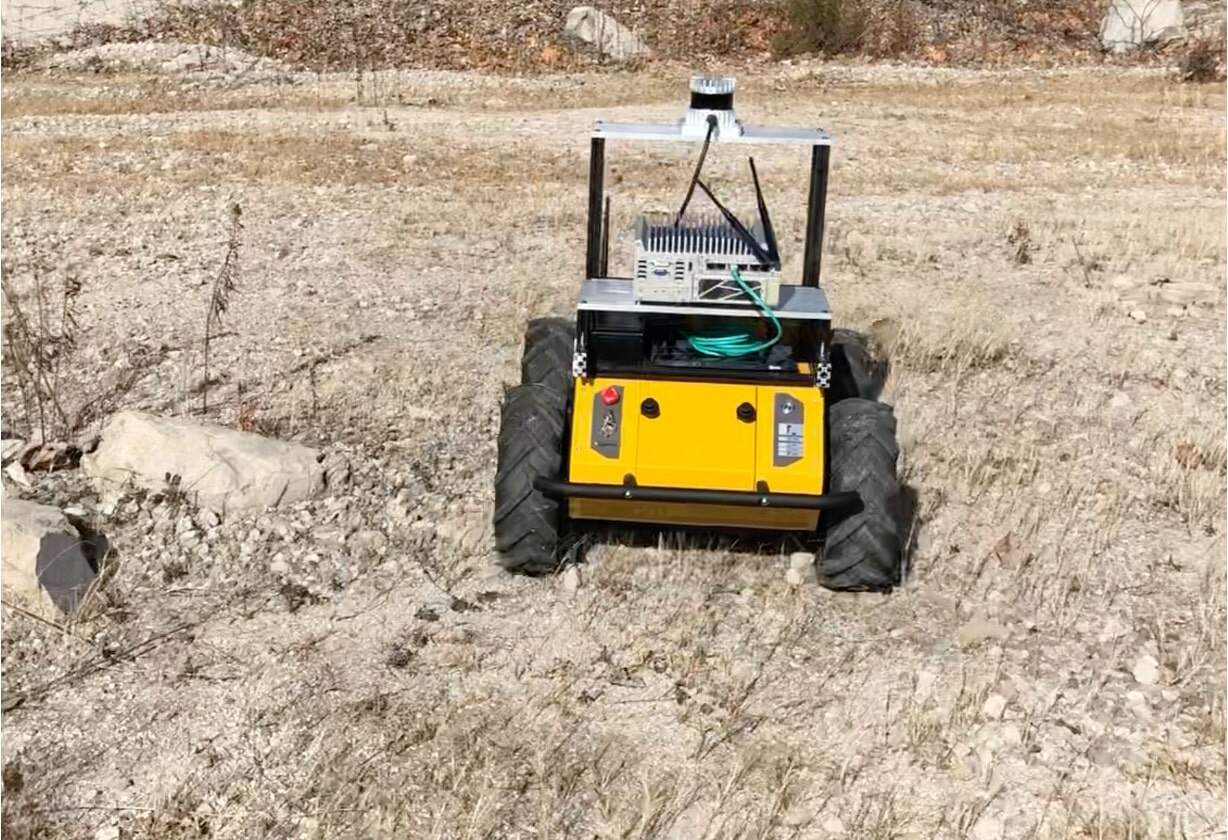}}\\
    \subfloat{\includegraphics[width=0.24\linewidth,height=1.1in]{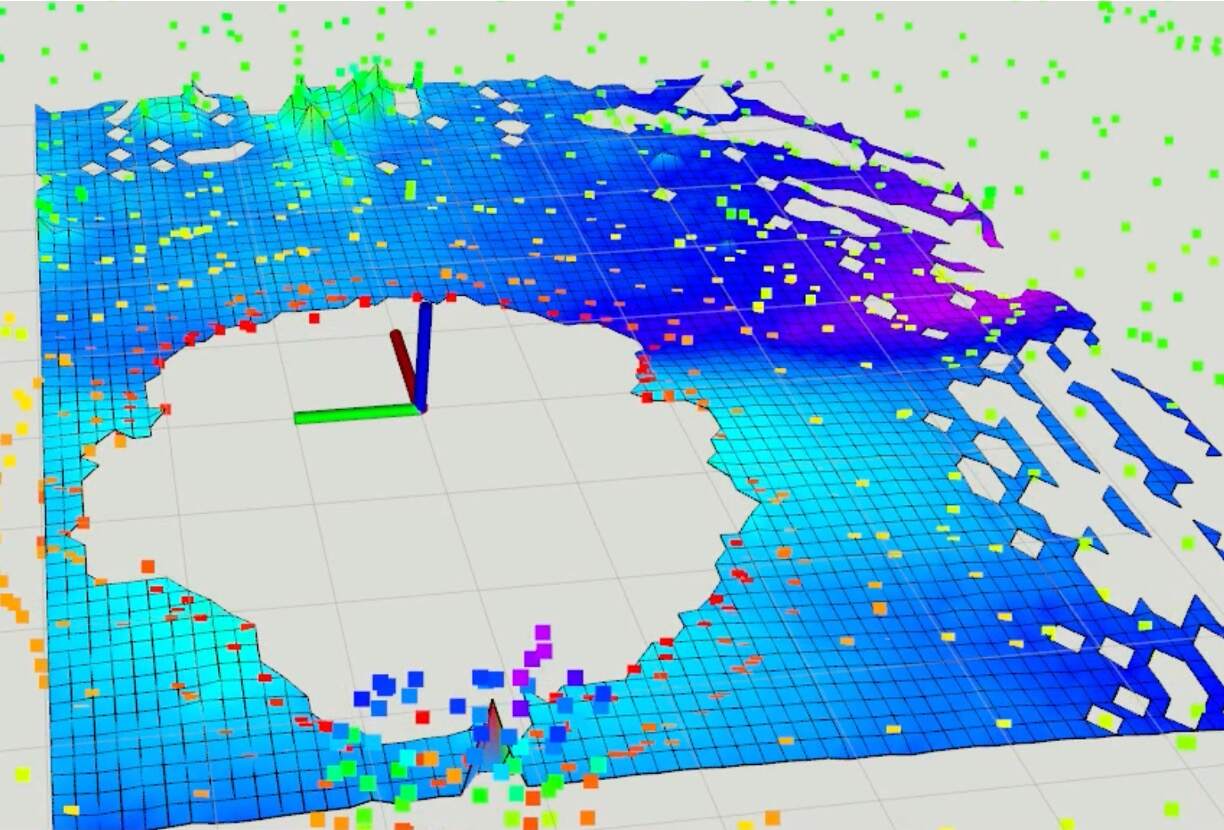}}
    \hspace{0.1cm}
    \subfloat{\includegraphics[width=0.24\linewidth,height=1.1in]{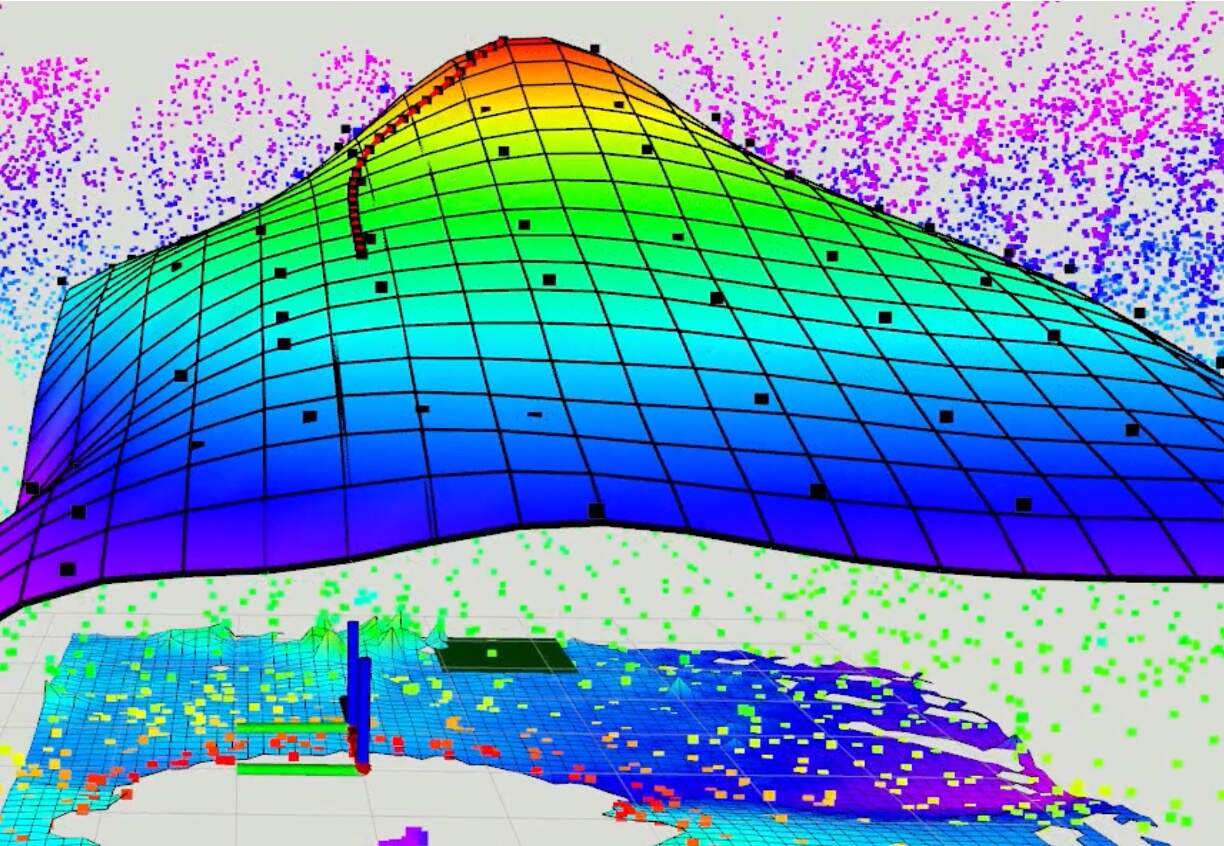}}
    \hspace{0.1cm}
    \subfloat{\includegraphics[width=0.24\linewidth,height=1.1in]{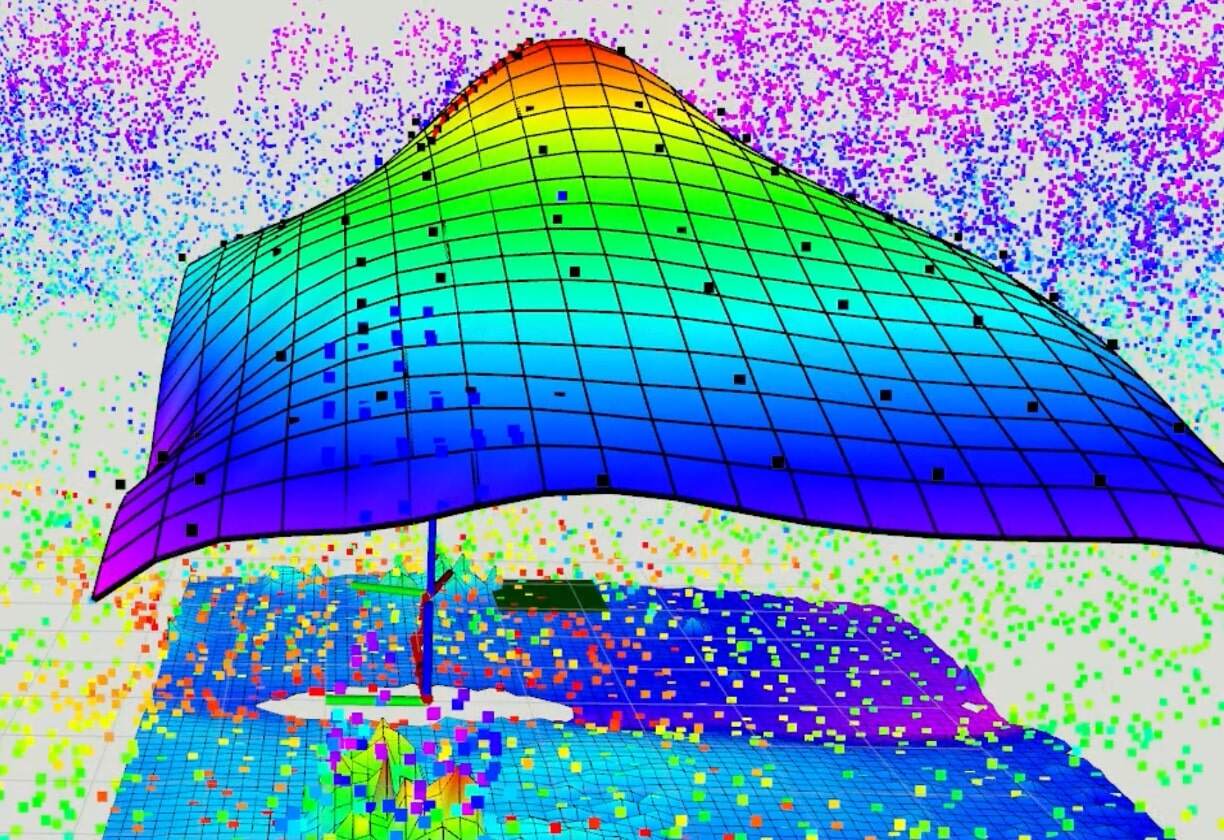}}
    \hspace{0.1cm}
    \subfloat{\includegraphics[width=0.24\linewidth,height=1.1in]{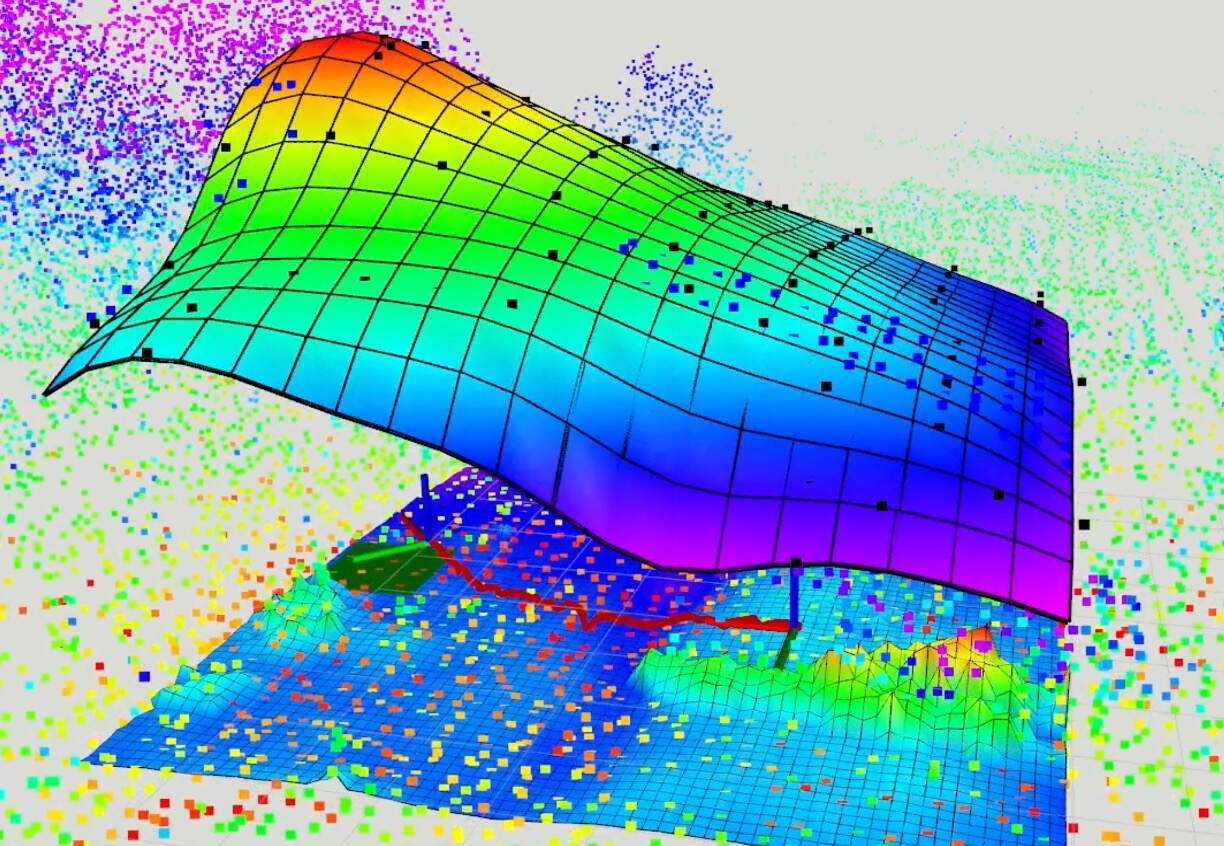}}\\
    \subfloat{\includegraphics[width=0.24\linewidth,height=1.1in]{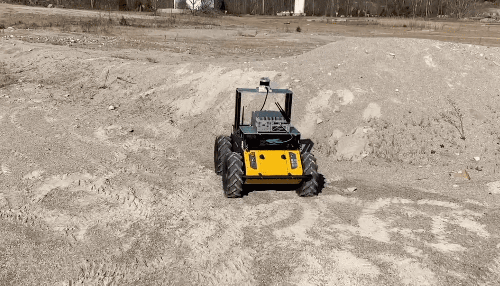}}
    \hspace{0.1cm}
    \subfloat{\includegraphics[width=0.24\linewidth,height=1.1in]{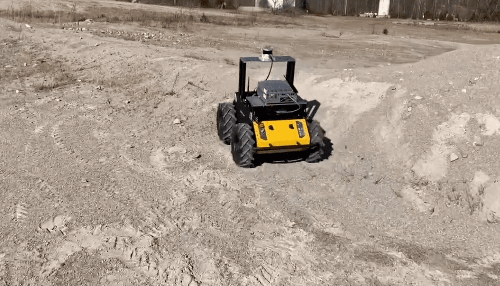}}
    \hspace{0.1cm}
    \subfloat{\includegraphics[width=0.24\linewidth,height=1.1in]{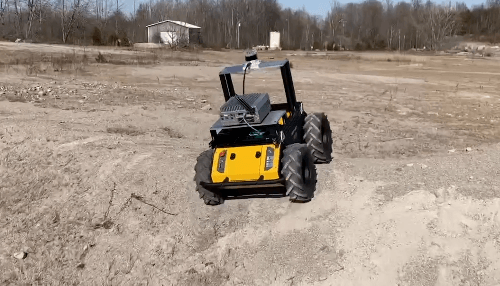}}
    \hspace{0.1cm}
    \subfloat{\includegraphics[width=0.24\linewidth,height=1.1in]{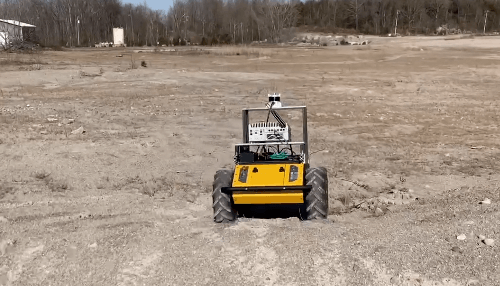}}\\
    \subfloat{\includegraphics[width=0.24\linewidth,height=1.1in]{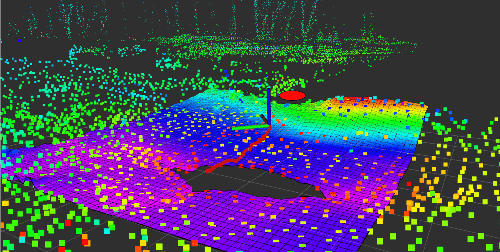}}
    \hspace{0.1cm}
    \subfloat{\includegraphics[width=0.24\linewidth,height=1.1in]{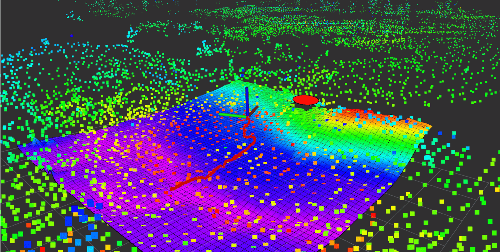}}
    \hspace{0.1cm}
    \subfloat{\includegraphics[width=0.24\linewidth,height=1.1in]{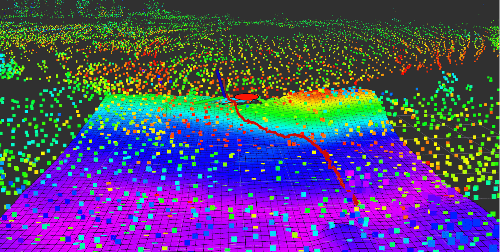}}
    \hspace{0.1cm}
    \subfloat{\includegraphics[width=0.24\linewidth,height=1.1in]{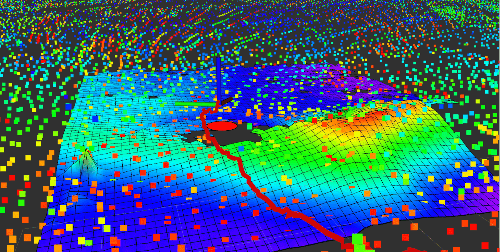}}\\
    \caption{ 
    Snapshots of navigation in two elevation-rich environments. 
    The first and second rows show the robot's planning in an environment scattered with rocks and the corresponding visualization for computing the solution. 
    In each column, the visualization showcases the elevation map and the value function, represented as the lower and upper surfaces, respectively. 
    Supporting states are marked with black dots for low-slope regions and blue dots for high-slope regions. 
    High-slope regions present a greater navigation risk for the robot and should be avoided. 
    The target destination for the robot is marked by a dark-green square.
    The red trajectory in the last column demonstrates the robot's traversed path. 
    The last two rows demonstrate the planning over an area containing a small hill and the corresponding visualization. 
    In the visualization, the red trajectory is the robot's traversed path and the red circle denotes the robot's planning goal.
    The color-coded surface represents the elevation map (for better visualization, the value function surface is not shown in this scenario).
    }
    \label{fig:elevation-rich-exp}
\end{figure*}

We further test our system in two outdoor environments shown in Fig.~\ref{fig:elevation-rich-exp} and Fig.~\ref{fig:trail}.
These environments contain unstructured terrains and irregular obstacles that introduce not only challenges in generating the obstacle-avoidance behavior but also difficulties in maintaining the robot's stable motions while traversing uneven/rough surfaces. 

\subsubsection{Navigation on Uneven Terrains:}

Fig.~\ref{fig:elevation-rich-exp} illustrates the robot's performance of navigating outdoor environments with rich elevation variations representing typical characteristics of uneven/rough terrains.

The first row of Fig.~\ref{fig:elevation-rich-exp} displays the navigation process over a complex terrain containing rocks, gravel, and short shrubs. 
To the robot, the rocks and short shrubs are impassable obstacles that are often difficult to map, and the gravel can disturb motions. 
The second row of the figure offers a detailed view of the robot's behavior guided by the computed value function. 
In this experiment, we demonstrate that the robot can effectively detect high-risk zones characterized by rocks and shrubs, even when relying solely on the geometric information generated by the elevation map. 
These high-risk zones are defined as the terrains with slopes larger than $45^{\circ}$.
We apply importance sampling to cover these zones in the corresponding state space during planning. 
Consequently, a dense cluster of states, indicated by blue dots, primarily encompasses the high-risk zones. 
While the states in high-risk regions provide information about infeasibility of traversal, others in low-slope regions offer insights into varying degrees of navigability, depending on the magnitude of the slope.
From the figure, we can see that the computed value function effectively integrates the environmental geometric information, resulting in a state-value surface with lower values over the high-slope areas. 
The low-valued portion of the surface reflects the increased costs of navigating the corresponding region. Thus, the associated policy 
steers the robot away from such hazardous areas.
Additionally, because in this scenario the value function is computed over a relatively large area, the disturbance caused by the gravel terrain can be effectively reduced as long as the robot remains within the area. 
The final snapshot shows that the robot can successfully follow the policy to reach the goal region, represented by the value function's maximum, demonstrating the efficacy of our method in this representative terrain.

The third row of Fig.~\ref{fig:elevation-rich-exp} shows another environment that presents unique challenges with a small hill peaking at approximately $1m$ in height, a steep slope in the middle, and a mild slope situated to the left of the robot's initial position.
The state space is defined over a $10m^2$ elevation map, and the states are sampled using importance sampling, where the slope determines the weights. 
The robot's initial position is purposefully set to face the steepest part of the terrain, directly confronting the environment's most challenging portion.
We use the frontier-based method~\citep{yamauchi1997frontier} to select temporary goals.
As shown in Fig.~\ref{fig:elevation-rich-exp}, the temporary goal, indicated as the red circle, is placed in the middle of the frontier region, located close to the obstructed area by the steepest and highest terrain segments.
Such goal selection poses a challenge for the robot. 
Although the shortest path between the robot's initial position and the goal is a straight line across the steep terrain, the risk of flipping or tipping over is also the largest. 
Despite this challenge, as seen in the first three columns of Fig.~\ref{fig:elevation-rich-exp}, our system effectively guides the robot to perform safe maneuvers to circumvent the steep terrain. 
This result proves that our method can effectively guide the robot to navigate hazardous terrains by penalizing the robot's movement distance using the first moment and incorporating the motion uncertainty via the second moment.
The last column shows that after arriving at the current goal, the vehicle continues its mission by extending its elevation map, updating the MDP, and computing a new policy based on the goal selected by the frontier-based method.

\subsubsection{Navigation on an Unstructured Construction Site:}
\begin{figure*} \vspace{15pt}
    \subfloat{\includegraphics[width=0.24\linewidth,height=1.1in]{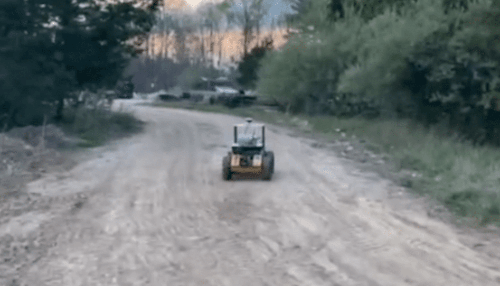}}
    \hspace{0.1cm}
    \subfloat{\includegraphics[width=0.24\linewidth,height=1.1in]{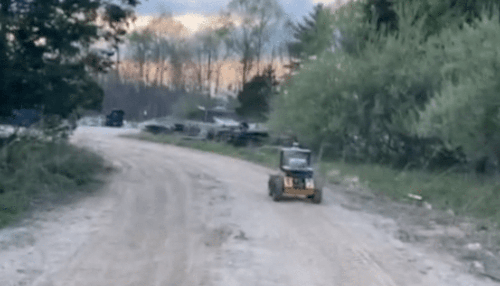}}
    \hspace{0.1cm}
    \subfloat{\includegraphics[width=0.24\linewidth,height=1.1in]{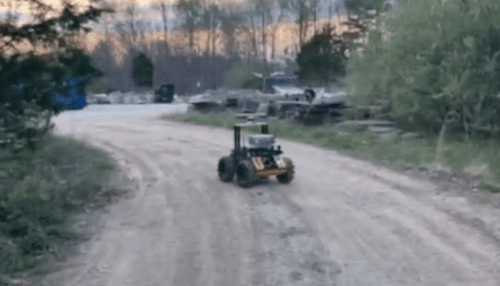}}
    \hspace{0.1cm}
    \subfloat{\includegraphics[width=0.24\linewidth,height=1.1in]{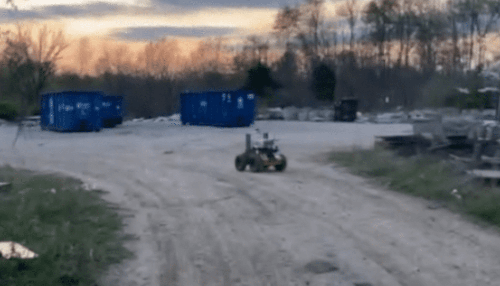}}\\
    \subfloat{\includegraphics[width=0.24\linewidth,height=1.1in]{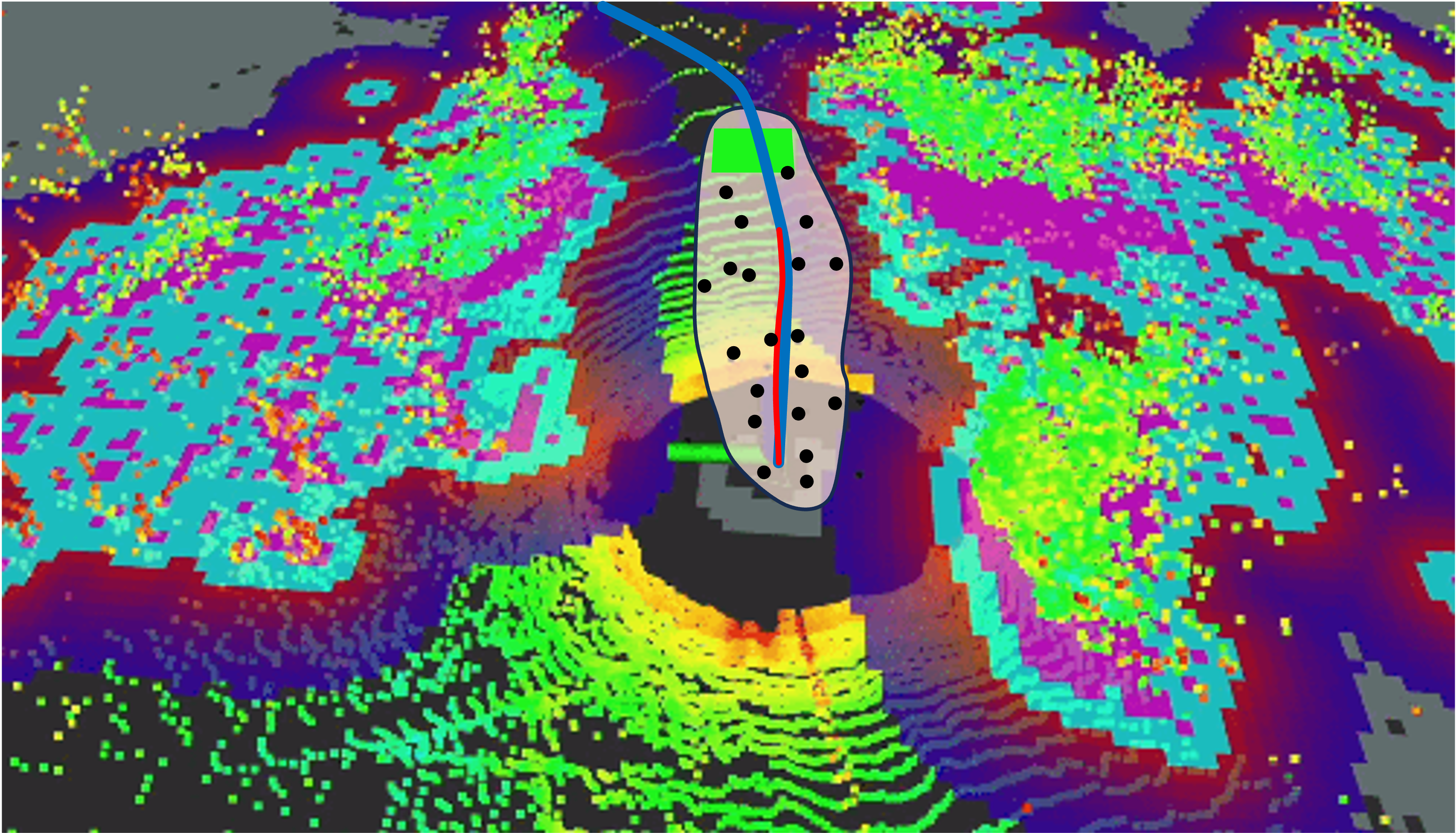}}
    \hspace{0.1cm}
    \subfloat{\includegraphics[width=0.24\linewidth,height=1.1in]{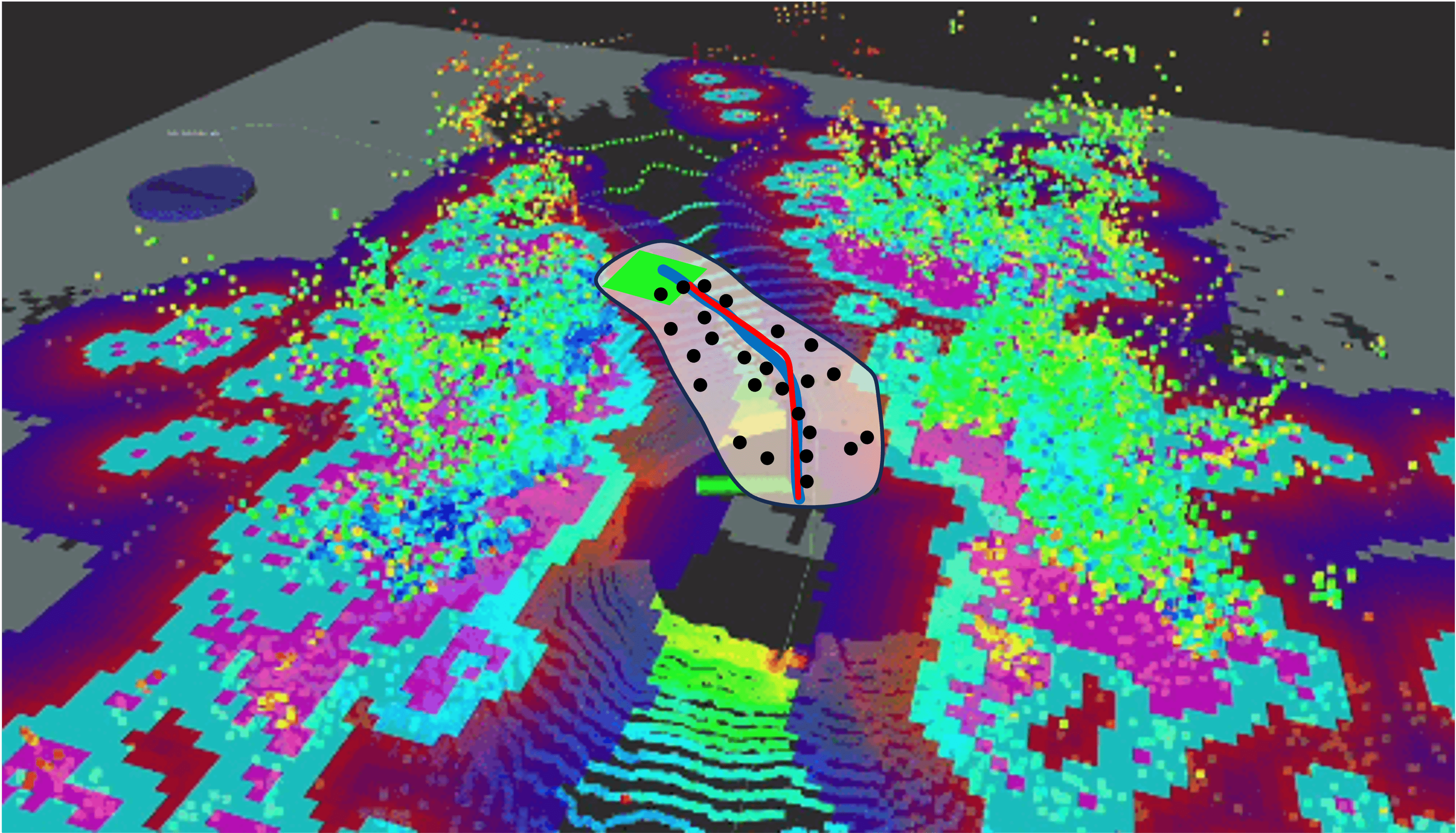}}
    \hspace{0.1cm}
    \subfloat{\includegraphics[width=0.24\linewidth,height=1.1in]{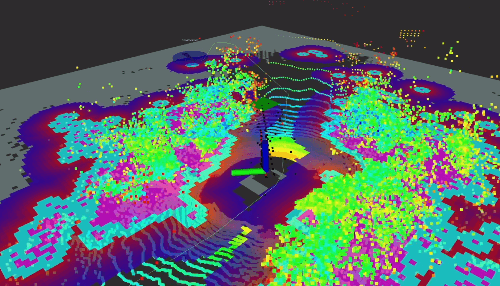}}
    \hspace{0.1cm}
    \subfloat{\includegraphics[width=0.24\linewidth,height=1.1in]{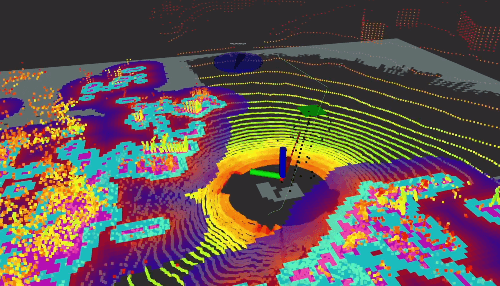}}
    \caption
    { 
    Snapshots of navigation in a trail using trajectory sampling to sample the states and cost map to represent the environment. 
    The visualization in the second row shows the cost map as a $2D$ color map, where a darker color means a higher cost. 
    In the first two visualizations, thickened/emphasized annotations are overlaid on top of the cost map. 
    Specifically, the red and blue trajectories represent the expected path by executing the policy and the global path, respectively.
    The scattered black dots adjacent to the global trajectory represent a selected subset of sampled states, included specifically for illustrative purposes. 
    Furthermore, the region encapsulating these states signifies the effective domain of the policy. 
    This is defined as the area within which the value function yields a positive value.
    The goal for each time step is shown as a green square.
    The final goal is the blue circle.
    }
    \label{fig:trail}
\end{figure*}

In contrast to the elevated environment, Fig.~\ref{fig:trail} shows the robot navigating a trail surrounded by dense vegetation around a construction site. 
Though lacking the challenges of steep slopes, this environment introduces new complexities. 
Particularly, dense vegetation makes up irregular and cluttered obstacles, and the unpaved road causes significant motion disturbance due to small rocks.

In order to adapt to this environment, we leverage a cost map to assign obstacle penalties. 
Due to the absence of elevation we employ the trajectory sampling mode with a global planner (ARA*).
The global planner is designed for obstacle avoidance on flat surfaces.
We set the temporary goal on the global path segment $6m$ ahead of the robot.
Because the global planner does not account for the uncertainty introduced by the unpaved trail, following the global path without considering these uncertainties may lead to large deviations and potential failures.
Our method can naturally handle this problem as shown in Fig.~\ref{fig:trail}. This figure demonstrates the robot's behavior at {4} timesteps.
The first two columns of the second row provide visualization of the planning process.
Specifically, 
the black dots represent the sampled supporting states around the blue global trajectory, closely followed by the expected robot's trajectory computed by our method, shown in red. 
In contrast to the deterministic trajectory optimization, which provides only a single trajectory, our method provides a feedback policy in a local space, represented as the half-transparent irregular shape encapsulating the supporting states.
This policy region, defined by the spread of the supporting states and the second moment of the stochastic motion, enables the robot to remain within the global-path-guided region while navigating toward the goal.
Thus, the system remains robust even when small rocks or other minor obstacles perturb the robot away from the globally planned path.
This capability enhances its resilience to typical disturbances of unpaved paths and is critical for the safety of our robot.

\section{Conclusion and Discussion}\label{sec:conclusion}
This paper presents 
a new decision making framework for robot planning and control in complex and unstructured environments such as the off-road navigation. 
We propose a method to solve the continuous-state Markov Decision Process by integrating the kernel value function representation and the Taylor-based approximation to Bellman optimality equation. 
Our algorithm alleviates the need for heavily searching in continuous state space and the need for precisely modeling the state transition functions. 
We have validated  the proposed method through thorough evaluations in both simplified and realistic planning scenarios.  
The experiments comparing with other baseline approaches show that our proposed framework is powerful and flexible,  and the performance statistics reveal superior efficiency and accuracy of the presented algorithm. 

In addition to the theoretical contribution, our real-world experiments reveal several challenges of applying the proposed method in practice. 
First, 
the precision of the optimal solution is contingent upon the specific locations of the supporting states within the kernel representation of the value function. Identifying the most suitable locations for these states is an important aspect that impacts the accuracy of the solution. One of our future focused areas will be designing techniques to determine these optimal state locations such as leveraging more advanced kernels as well as utilizing more advanced hyperparameter learning schemes. 
Moreover, we aim to extend this investigation into the scalability of our approach within high-dimensional spaces. Addressing the challenges associated with scalability in the expansive spaces stands as an important objective for our forthcoming research efforts. 
Second, we present a system (Section~\ref{sec:system}) which is mainly responsible for converting the raw sensor data to an MDP problem by assuming that feasible/infeasible regions can be distinguished and the features in the environment (elevation in our experiment) can be obtained accurately from sensors. 
Since this MDP is used for computing a policy, ensuring the MDP matches the real-world scenario is paramount in the final performance of the system. 
However, the map built from the noisy sensor measurement usually cannot accurately represent the geometry of the environment, e.g., occupancy and elevation. 
The MDP derived from this inaccurate map may deviate from the real environment. 
Thus, reasoning about these inaccuracies in the planning method is essential for building a robust navigation system. 
Additionally, an accurate physics model for planning is necessary in complex environments, where detailed motion control strategies are needed.
By comparing the results of the Section~\ref{sec:first-exp} and Section~\ref{sec:real-world-exp}, we can observe that the planning method can generate more efficient policies if we use a physics model which can better describe the robot's motion. 
In our real-world experiments, our modeling of the first moment uses inaccurate models to describe the physical interaction between the robot and the terrain.
Although these models can be used for planning in the particular environments we tested, to generalize to more complex situations,  it is necessary to develop more accurate physical models that consider not only the elevation but also terrain textures, which is our future work. 

\section*{Funding} 
The authors disclosed receipt of the following financial support for the research, authorship, and/or publication of this article. 
This work has been supported by the Army Research Office and was accomplished under Cooperative Agreement Numbers W911NF-20-2-0099 and W911NF-22-2-0018: Scalable, Adaptive, and Resilient Autonomy (SARA).  The views and conclusions contained in this document are those of the authors and should not be interpreted as representing the official policies, either expressed or implied, of the Army Research Office or the U.S. Government. The U.S. Government is authorized to reproduce and distribute reprints for Government purposes notwithstanding any copyright notation herein.
This work was also partially supported by NSF: RI: Small: Exploiting Symmetries of Decision Theoretic Planning for Autonomous Vehicles (grant no 2006886), and NSF: CAREER: Autonomous Live Sketching of Dynamic Environments by Exploiting Spatiotemporal Variations (grant no 2047169).


\end{document}